\documentclass[10pt,twocolumn,letterpaper]{article}

\usepackage{cvpr}
\usepackage{times}
\usepackage{epsfig}
\usepackage{graphicx}
\usepackage{amsmath}
\usepackage{amssymb}

\usepackage{amsthm}
\usepackage{subfig}
\usepackage{tabularx}
\usepackage{booktabs}
\usepackage{multirow}
\usepackage{multicol}
\usepackage{enumitem}
\usepackage{array}
\usepackage{color, colortbl}
\usepackage{fixltx2e}

\definecolor{Gray}{gray}{0.92}

\definecolor{sh_gray}{rgb}{0.84,0.84,0.84}
\definecolor{sh_gray2}{rgb}{1,0.89,0.75}
\definecolor{sh_gray3}{rgb}{1,0.7,0.6}

\newcommand{\bsigma}{\boldsymbol{\sigma}}
\newcommand{\bmu}{\boldsymbol{\mu}}
\newcommand{\btheta}{\boldsymbol{\theta}}

\newcommand{\bx}{\mathbf{x}}

\newcommand{\bz}{\mathbf{z}}
\newcommand{\bs}{\mathbf{s}}
\newcommand{\bw}{\mathbf{w}}
\newcommand{\br}{\mathbf{r}}
\newcommand{\worst}[1]{\colorbox{sh_gray3}{\textbf{#1}}}%
\newcommand{\worse}[1]{\colorbox{sh_gray2}{\textbf{#1}}}%
\newcommand{\better}[1]{\colorbox{sh_gray}{\textbf{#1}}}%

\usepackage[breaklinks=true,bookmarks=false]{hyperref}

\cvprfinalcopy % *** Uncomment this line for the final submission

\def\cvprPaperID{4510} % *** Enter the CVPR Paper ID here

\ifcvprfinal\fi
\begin{document}

%%%%%%%%% TITLE
\title{Data Uncertainty Learning in Face Recognition}

\author{Jie Chang$^1$, Zhonghao Lan$^2$, Changmao Cheng$^1$, Yichen Wei$^1$\\
{$^1$Megvii Inc.}
{$^2$University of Science and Technology of China}\\
{\texttt{\small{\{changjie, lanzhonghao, chengchangmao, weiyicheng\}@megvii.com}} \hspace{0.5cm}}\\
}

\maketitle
%\thispagestyle{empty}

%%%%%%%%% ABSTRACT
\begin{abstract}
Modeling data uncertainty is important for noisy images, but seldom explored for face recognition. The pioneer work~\cite{shi2019probabilistic} considers uncertainty by modeling each face image embedding as a Gaussian distribution. It is quite effective. However, it uses fixed feature (mean of the Gaussian) from an existing model. It only estimates the variance and relies on an ad-hoc and costly metric. Thus, it is not easy to use. It is unclear how uncertainty affects feature learning.

This work applies data uncertainty learning to face recognition, such that \textbf{the feature (mean) and uncertainty (variance) are learnt simultaneously}, for the first time. Two learning methods are proposed. They are easy to use and outperform existing deterministic methods as well as~\cite{shi2019probabilistic} on challenging unconstrained scenarios. We also provide insightful analysis on how incorporating uncertainty estimation helps reducing the adverse effects of noisy samples and affects the feature learning.
\end{abstract}

%%%%%%%%% BODY TEXT
\section{Introduction}
\textit{Data uncertainty}\footnote{\textit{Uncertainty} could be characterised into two main categories. Another type is \textit{model uncertainty}.} captures the ``noise'' inherent in the data. Modeling such uncertainty is important for computer vision application~\cite{kendall2017uncertainties}, e.g., face recognition, because noise widely exists in images. 

\begin{figure}[t]
    \centering
    \captionsetup{font=footnotesize}
    \vspace{-0.5em}\includegraphics[width=1.0\linewidth]{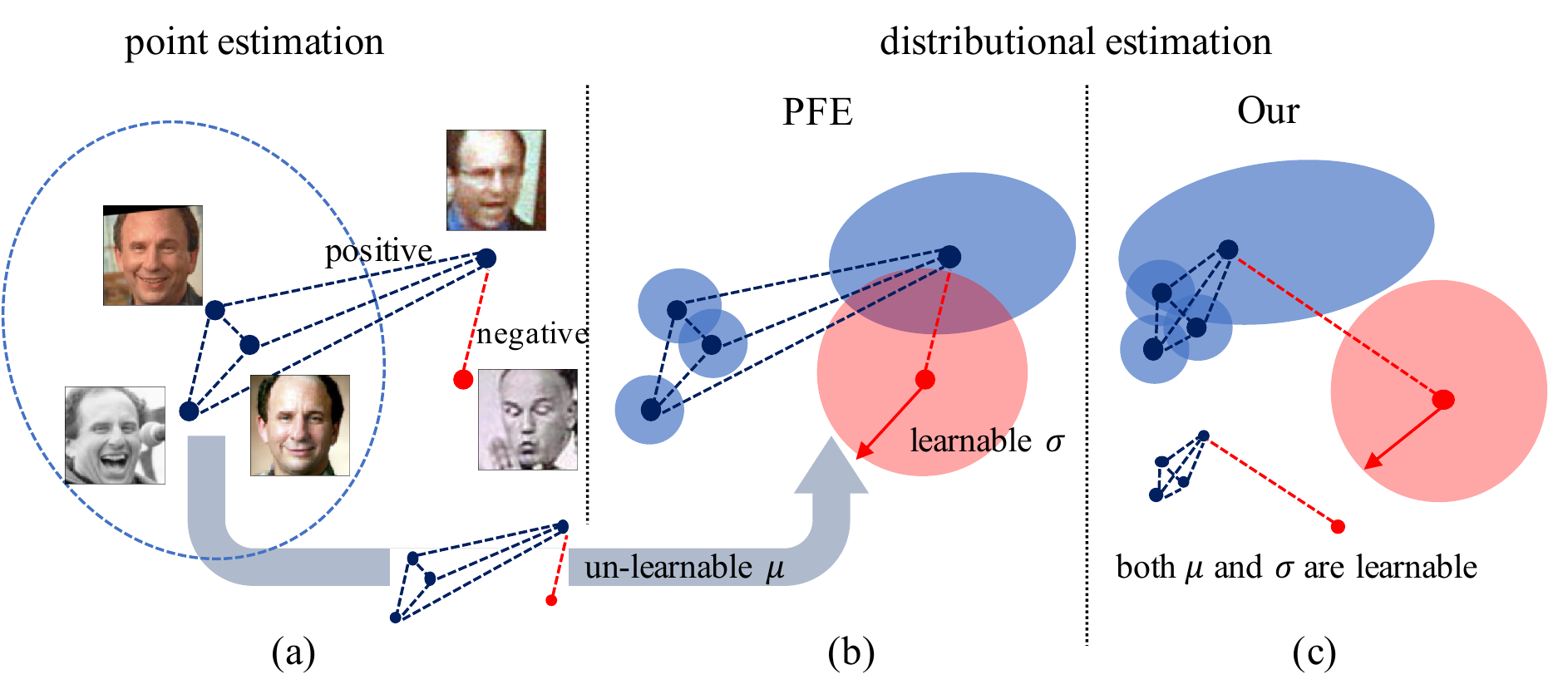}\\[-1.0em]
    \vspace{-0.2em}\caption{(a) Deterministic model gives point embedding without considering the data uncertainty; (b) probabilistic model gives distributional estimation parameterised with estimated mean and estimated variance. PFE leverages the pre-trained point embedding as the mean $\bmu$, \textbf{only} learn the uncertainty $\bsigma$ for each sample; (c) our method simultaneously learn $\bsigma$ as well as $\bmu$, leading to better intra-class compactness and inter-class separability for $\bmu$ in the latent space. Different classes are marked as blue or red. \textbf{Best viewed in color.}}\vspace{-1.5em}
    \label{fig:embedding}
\end{figure}

Most face recognition methods represent each face image as a deterministic point embedding in the latent space~\cite{deng2019arcface,liu2017sphereface,wang2018cosface,wen2016discriminative,schroff2015facenet}. Usually, high-quality images of the same ID are clustered. However, it is difficult to estimate an accurate point embedding for noisy face images, which are usually out of the cluster and have larger uncertainty in the embedding space. This is exemplified in Fig~\ref{fig:embedding} (a). The positive example is far from its class and close to a noisy negative example, causing a mismatch. 

Probabilistic face embeddings (PFE)~\cite{shi2019probabilistic} is the first work to consider \textit{data uncertainty} in face recognition. For each sample, it estimates a Gaussian distribution, instead of a fixed point, in the latent space. Specifically, given a pre-trained FR model, \emph{the mean of the Gaussian for each sample is fixed as the embedding produced by the FR model}. An extra branch is appended to the FR model and trained to estimate the variance. The training is driven by a new similarity metric, mutual likelihood score or \textit{MLS}, which measures the ``likelihood'' between two Gaussian distributions. It is shown that PFE estimates small variance for high-quality samples but large variance for noisy ones. Together with the \emph{MLS} metric, PFE can reduce the mismatches on noisy samples. This is illustrated in Fig~\ref{fig:embedding}, (b). While being effective, PFE is limited in that \emph{it does not learn the embedded feature (mean) but only the uncertainty}. As a result, it is unclear how uncertainty affects feature learning. Also, the conventional similarity metric such as cosine distance cannot be used. The more complex \emph{MLS} metric is in demand, which takes more runtime and memory.

For the first time, this work applies \textit{data uncertainty learning} (DUL) to face recognition such that \emph{feature (mean) and uncertainty (variance) are learnt simultaneously}. As illustrated in Fig~\ref{fig:embedding} (c), this improves the features such that the instances in the same class are more compact and the instances in different classes are more separated. In this case, the learned feature is directly usable for conventional similarity metric. \emph{MLS} metric is no longer necessary. 

Specifically, we propose two learning methods. The first is classification based. It learns a model from scratch. The second is regression based. It improves an existing model , similar as PFE. We discuss how the learned uncertainty affects the model training in two methods, from the perspective of \textit{image noise}. We provide insightful analysis that the learned uncertainty will improve the learning of identity embeddings by \emph{adaptively reducing the adverse effects of noisy training samples}.

Comprehensive experiments demonstrate that our proposed methods improve face recognition performance over existing deterministic models and PFE on most public benchmarks. The improvement is more remarkable on benchmarks with low quality face images, indicating that model with \textit{data uncertainty learning} is more suitable to unconstrained face recognition scenario, thus important for practical tasks.

\section{Related Work}
\label{sec:related-work}

\paragraph{Uncertainty in Deep Learning} 

The nature of uncertainties as well as the manner to deal with them have been extensively studied to help solve the reliability assessment and risk-based decision making problems for a long time~\cite{faber2005treatment,pate1996uncertainties,der2009aleatory}.
In recent years, uncertainty is getting more attention in deep learning. Many techniques have been proposed to investigate how uncertainty specifically behaves in deep neural networks~\cite{blundell2015weight,gal2016uncertainty,gal2016dropout,kendall2017uncertainties}. 
Specific to deep uncertainty learning, \textit{uncertainties} can be be categorised into \textit{model uncertainty} capturing the noise of the parameters in deep neural networks, and \textit{data uncertainty} measuring the noise inherent in given training data. 
Recently, many computer vision tasks, i.e., semantic segmentation~\cite{isobe2017deep,kendall2015bayesian}, object detection~\cite{Choi_2019_ICCV,kraus2019uncertainty} and person Re-ID~\cite{yu2019robust}, have introduced deep uncertainty learning to CNNs for the improvement of model robustness and interpretability. 
In face recognition task, several works have been proposed to leverage \textit{model uncertainty} for analysis and learning of face representations~\cite{gong2017capacity,zafar2019face,khan2019striking}. Thereinto PFE~\cite{shi2019probabilistic}, is the first work to consider \textit{data uncertainty} in face recognition task. 

\paragraph{Noisy Data Training} Large-scale datasets, i.e., CASIA-WebFace~\cite{yi2014learning}, Vggface2~\cite{cao2018vggface2} and MS-Celeb-1M~\cite{guo2016ms}, play the important role in training deep CNNs for face recognition. It is inevitable these face datasets collected online have lots of \textit{label noise} --- examples have erroneously been given the labels of other classes within the dataset. Some works explore the influence of \textit{label noise}~\cite{wang2018devil} and how to train robust FR models in this case~\cite{hu2019noise,wu2018light,ng2014data}. Yu~\etal~\cite{yu2019robust} claims in person Re-ID that another \textit{image noise} brought by poor quality images also has detrimental effect on the trained model. Our methods are not specifically proposed for noisy data training, however, we provide insigtful analysis about how the learned data uncertainty affect the model training from the perspective of \textit{image noise}. Additionally, we experimentally demonstrate the proposed methods perform more robustly on noisy dataset.

\section{Methodology}
\label{sec:Methodology}
In Section~\ref{sec:preliminaries}, we first reveals the data uncertainty inherently existed in continuous mapping space and our specific face datasets. In Section~\ref{sec:classification}, we propose DUL\textsubscript{cls} to consider \textit{data uncertainty learning} in a standard face classification model. We next propose another regression-based method, DUL\textsubscript{rgs} to improve existing deterministic models in Section~\ref{sec:regression}. 
Last in Section~\ref{sec:discussion}, we clarify some differences between proposed methods and existing works.

\subsection{Preliminaries}
\label{sec:preliminaries}

\begin{figure}[t]
    \centering
    \captionsetup{font=footnotesize}
    \begin{minipage}{0.48\linewidth}
    \includegraphics[width=1.0\linewidth]{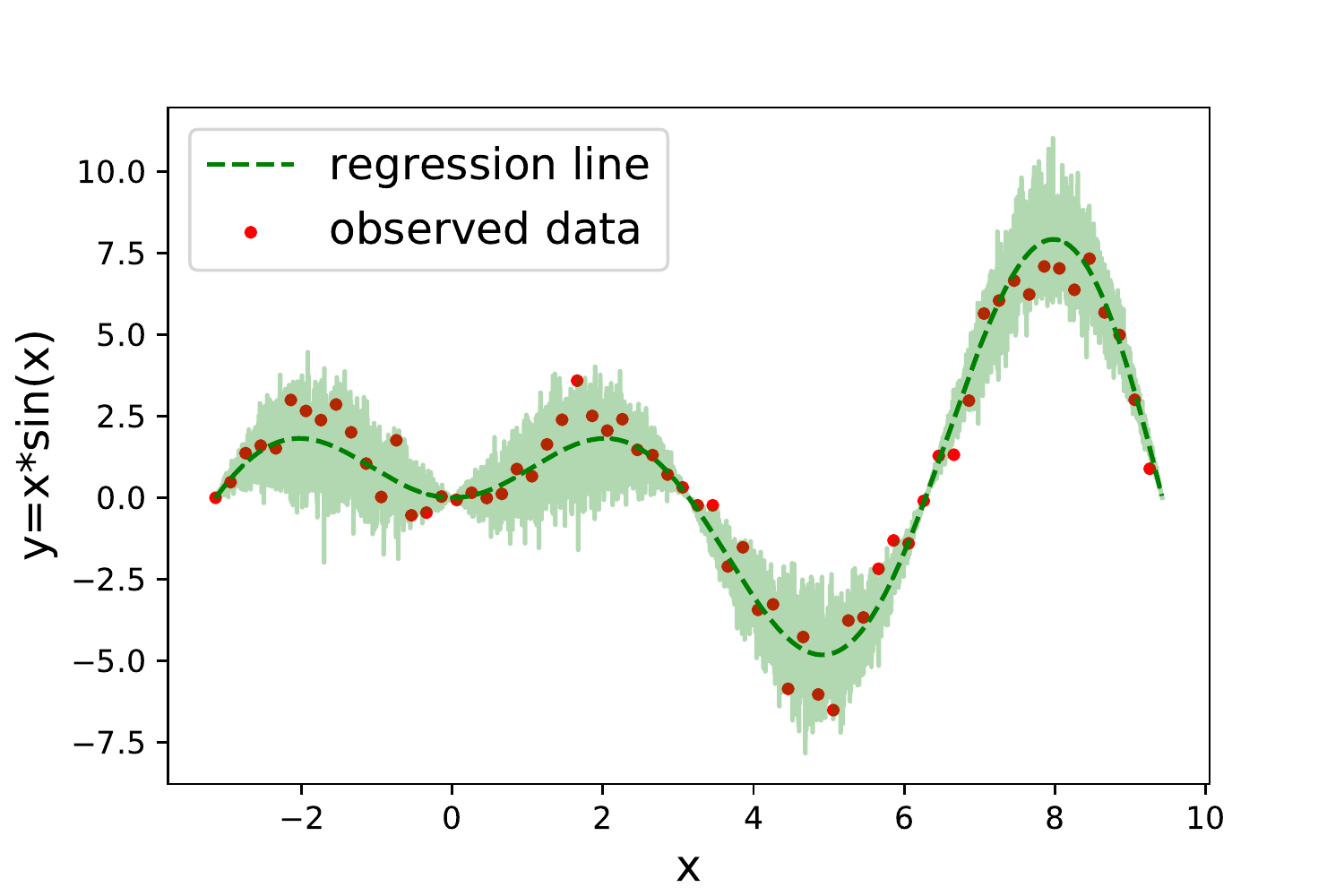}\\[-2.0em]
    \begin{center} \footnotesize(a) noise in mapping $\mathcal{X}\rightarrow\mathcal{Y}$.  \end{center}\vspace{-1.2em}
    \end{minipage}\hfill
    \begin{minipage}{0.48\linewidth}
    \includegraphics[width=1.0\linewidth]{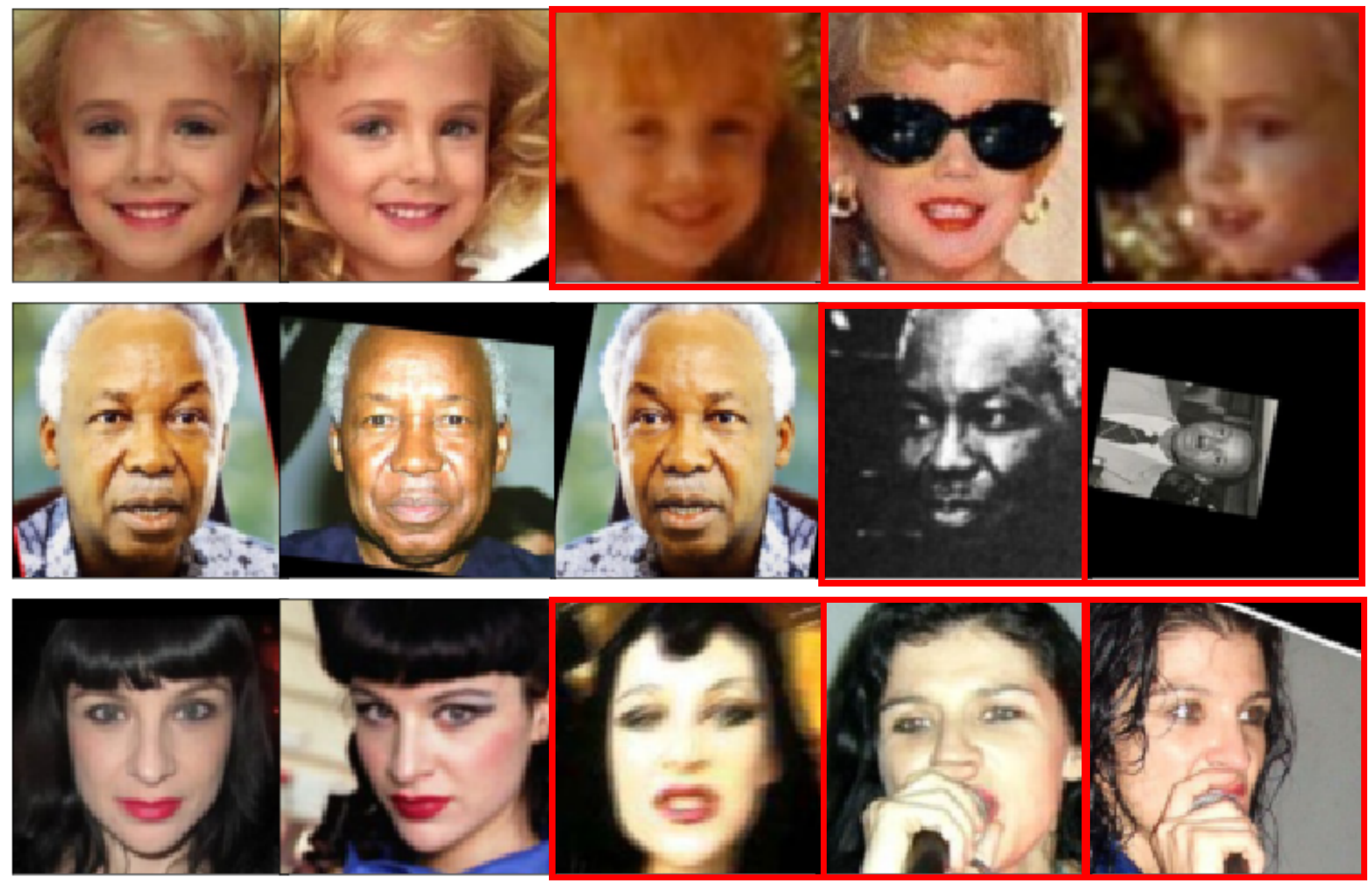}\\[-2.0em]
    \begin{center} \footnotesize(b) ``noise'' in MS-Celeb-1M. \end{center}\vspace{-1.2em}
    \end{minipage}\hfill
    \vspace{-0.3em}\caption{(a): Target $y$ in observed data pair (red-dot) is corrupted by $x$ dependent noise. \textit{Data uncertainty regression} will give us the ``noise level'' (green-shaded) beyond the particular predicted value (green line); (b): Samples labeled with the same ID are presented in each row. Samples with red box are regarded as noisy data compared with other intra-class samples. \textbf{Best viewed in color.}}
    \label{fig:regression}
\end{figure}

\paragraph{Uncertainty in Continuous Mapping Space}
Supposing a continuous mapping space $\mathcal{X}\rightarrow\mathcal{Y}$ where each $y_i\in \mathcal{Y}$ is corrupted by some input-dependent noise, $n(\bx_i), \bx_i\in\mathcal{X}$, then we say this mapping space carries \textit{data uncertainty} in itself.
Considering a simple case, the noise is additive and drawn from Gaussian distribution with mean of zero and $x$-dependent variance. Then each observation target $y_i = f(\bx_i) + \epsilon\sigma(\bx_i)$, where $\epsilon\sim\mathcal{N}(0,\mathbf{I})$ and $f(\cdot)$ is the embedding function we want to find. 
Conventional regression model only trained to approximate $f(\bx_i)$ given the input $\bx_i$. However, regression model with \textit{data uncertainty learning} also estimates $\sigma(\bx_i)$, representing the uncertainty of the predicted value $f(\bx_i)$ (see Fig~\ref{fig:regression}, (a)). This technique has been used by many tasks~\cite{kendall2017uncertainties,brando2018uncertainty,nix1994estimating,goldberg1998regression,bishop1997regression}.

\paragraph{Uncertainty in Face Datasets}
Similar to the above continues mapping space, face datasets composed with $\mathcal{X}\rightarrow Y$ also carries \textit{data uncerainty}. Here $\mathcal{X}$ is the continues image space while $Y$ is the discrete identity labels. Typically, large amount of face images collected online are visually ambiguous (poorly aligned, severely blurred or occluded). It is difficult to filter out these poor quality samples from training set (see Fig~\ref{fig:regression}, (b)). During deep learning era, each sample is represented as an embedding $\bz_i$ in the latent space.
If we hypothesize that each $\bx_i \in \mathcal{X}$ has an ideal embedding $f(\bx_i)$ mostly representing its identity and less unaffected by any identity irrelevant information in $\bx_i$, then the embedding predicted by DNNs can reformulated as $\bz_i = f(\bx_i) + n(\bx_i)$ where $n(\bx_i)$ is the uncertainty information of $\bx_i$ in the embedding space.

\subsection{Classification-based DUL for FR}
\label{sec:classification}
We propose DUL\textsubscript{cls} to firstly introduce \textit{data uncertainty learning} to the face classification model which can be trained end-to-end.

\begin{figure}[t]
    \centering
    \captionsetup{font=footnotesize}
    \vspace{-0.5em}\includegraphics[width=0.85\linewidth]{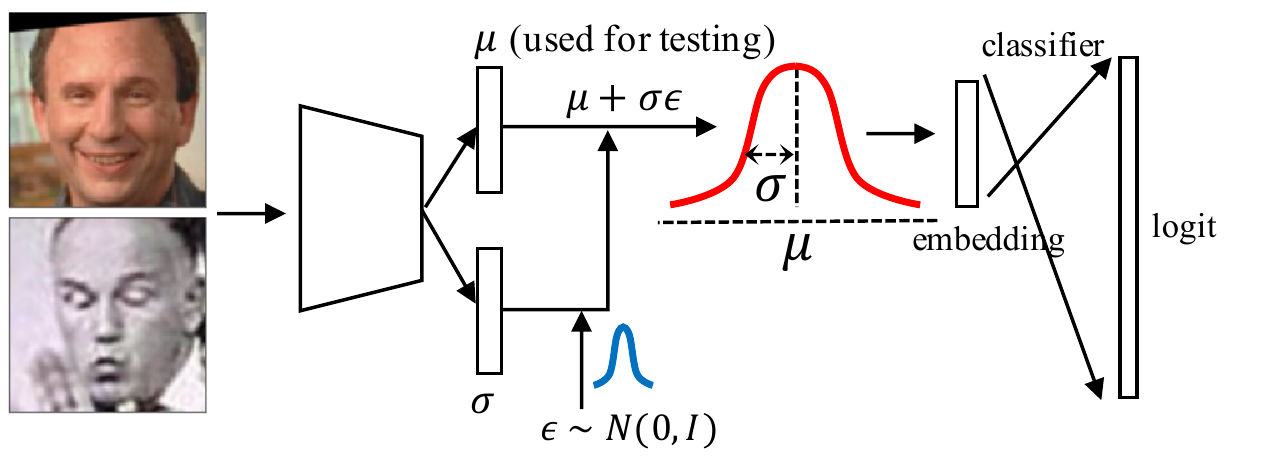}\\[-1.0em]
    \vspace{-0.2em}\caption{Overview of the proposed DUL\textsubscript{cls} FR model.}\vspace{-1.5em}
    \label{fig:dul-cls}
\end{figure}

\paragraph{Distributional Representation} 
Specifically, we define the representation $\bz_i$ in latent space of each sample $\bx_i$ as a Gaussian distribution,
\begin{equation}
p(\bz_i|\bx_i) = \mathcal{N}(\bz_i; \bmu_i, \bsigma_i^2\mathbf{I})
\label{eq:gaussian-distribution}
\end{equation}
where both the parameters (mean as well as variance) of the Gaussian distribution are input-dependent predicted by CNNs: $\bmu_i=f_{\btheta_1}(\bx_i)$, $\bsigma_i=f_{\btheta_2}(\bx_i)$, where $\btheta_1$ and $\btheta_2$ refer to the model parameters respectively w.r.t output $\bmu_i$ and $\bsigma_i$. 
Here we recall that the predicted Gaussian distribution is diagonal multivariate normal.
$\bmu_i$ can be regarded as the identity feature of the face and the $\bsigma_i$ refers to the uncertainty of the predicted $\bmu_i$. 
Now, the representation of each sample is not a deterministic point embedding any more, but a stochastic embedding \textit{sampled} from $\mathcal{N}(\bz_i; \bmu_i, \bsigma_i^2\mathbf{I})$, in the latent space. 
However, sampling operation is not differentiable preventing the backpropagation of the gradients flow during the model training. 
We use re-parameterization trick~\cite{kingma2013auto} to let the model still take gradients as usual.
Specifically, we first sample a random noise $\epsilon$ from a normal distribution, which is independent of the model parameters, and then generate $\bs_i$ as the equivalent sampling representation (see Fig~\ref{fig:dul-cls} for an overview pipeline), 
\begin{equation}
\bs_i = \bmu_i + \epsilon\bsigma_i, \quad \epsilon\sim \mathcal{N}(\mathbf{0},\mathbf{I}).
\label{eq:reparameterization}
\end{equation}

\paragraph{Classification Loss}
Since $\bs_i$ is the final representation of each image $\bx_i$, we feed it to a classifier to minimize the following softmax loss,
\begin{equation}
\mathcal{L}_{softmax} = \frac{1}{N}\sum_i^N -\log \frac{e^{\bw_{y_i}\bs_i}}{\sum_c^C e^{\bw_c\bs_i}},
\label{eq:softmax-loss}
\end{equation}
In practice, we use different variants of $\mathcal{L}_{softmax}$ such as additive margin~\cite{wang2018additive}, feature $\ell$2 normalization~\cite{ranjan2017l2} and arcface~\cite{deng2019arcface}, to train our face classification model.

\paragraph{KL-Divergence Regularization} 
Eq.~\ref{eq:reparameterization} indicates that all identity embeddings $\bmu_i$ are corrupted by $\bsigma_i$ during the training period, this will prompt the model to predict small $\bsigma$ for all samples in order to suppress the unstable ingredients in $\bs_i$ such that Eq.~\ref{eq:softmax-loss} can still converge at last.
In this case, the stochastic representation can be reformulated as $\bs_i = \bmu_i + c$ which is actually degraded to the original deterministic representation\footnote{Here $c$ refers to the estimated $\bsigma$ which nearly constant and small.}.
Inspired by the variational information bottleneck~\cite{alemi1612deep}, we introduce a regularization term during the optimization by explicitly constraining $\mathcal{N}(\bmu_i, \bsigma_i)$ to be close to a normal distribution, $\mathcal{N}(\mathbf{0},\mathbf{I})$, measured by Kullback-Leibler divergence (KLD) between these two distributions. This KLD term is, 
\begin{equation}
\begin{aligned}
\mathcal{L}_{kl} & = KL[N(\bz_i|\bmu_i, \bsigma_i^2)|| N(\epsilon|\mathbf{0}, \mathbf{I})] \\
& = -\frac{1}{2}(1+\log\bsigma^2-\bmu^2-\bsigma^2)
\end{aligned}
\label{eq:kl-loss}
\end{equation}
Noted that $\mathcal{L}_{kl}$ is monotonely decreasing w.r.t $\bsigma$ under the restriction that $\bsigma_i^{(l)} \in (0,1)$ ($l$ refers to the $l^{th}$ dimension of the embedding).
$\mathcal{L}_{kl}$ works as a good ``balancer'' with Eq.~\ref{eq:softmax-loss}. 
Specifically, DUL\textsubscript{cls} is discouraged from predicting large variance for all samples, which may lead to extremely corruption on $\bmu_i$, thus making $\mathcal{L}_{softmax}$ hard to converge.
Simultaneously, DUL\textsubscript{cls} is also discouraged from predicting lower variance for all samples, which may lead to larger $\mathcal{L}_{kl}$ to punish the model in turn.

Last, we use $\mathcal{L}_{cls} = \mathcal{L}_{softmax} + \lambda \mathcal{L}_{kl}$ as the total cost function, and $\lambda$ is a trade-off hyper-parameter, and it is further analysed in Section~\ref{sec: other-experiment}.

\subsection{Regression-based DUL for FR}
\label{sec:regression}
DUL\textsubscript{cls} is a general classification model with data uncertainty learning. 
Next we propose another regression based method, DUL\textsubscript{rgs}, improving existing FR models by data uncertainty learning. 

\paragraph{Difficulty of Introducing \textit{Data Uncertainty Regression} to FR}
DUL\textsubscript{rgs} is inspired from data uncertainty regression~\cite{le2005heteroscedastic,kendall2017uncertainties} for continuous mapping space $\mathcal{X}\rightarrow\mathcal{Y}$ as described in Section~\ref{sec:preliminaries}. However, mapping space in face datasets is constructed by continuous image space $\mathcal{X}$ and discrete identity label $\mathcal{Y}$, which cannot be directly fitted via data uncertainty regression. The key point lies in that the identity labels $y_c\in Y$ cannot serve as continues target vector to be approximated.
This difficulty is also mentioned in PFE~\cite{shi2019probabilistic} but is not resolved.

\paragraph{Constructing New Mapping Space for FR}
We construct a new target space, which is continuous, for face data. 
Most importantly, it is nearly equivalent to the original discrete target space $Y$, which encourages the correct mapping relationship. 
Specifically, we pre-train a classification-based deterministic FR model, and then utilize the weights in its classifier layers, $\mathcal{W}\in\mathbb{R}^{D\times C}$ as the expected target vector\footnote{Here $D$ refers to the dimensions of the embedding and $C$ refers to the numbers of classes in training set.}. 
Since each $\bw_i\in \mathcal{W}$ can be treated as the typical center of the embeddings with the same class, $\{\mathcal{X}, \mathcal{W}\}$ thus can be regarded as the new equivalent mapping sapce. 
Similar to the uncertainty in continuous mapping space as described in Section~\ref{sec:preliminaries}, $\{\mathcal{X}, \mathcal{W}\}$ has inherent noise. We can formulate the mapping from $\bx_i \in \mathcal{X}$ to $\bw_i\in \mathcal{W}$ as $\bw_i = f(\bx_i) + n(\bx_i)$, where $f(\bx_i)$ is the ``ideal'' identity feature and each observed $\bw_i$ is corrupted by input dependent noise.

\begin{figure}[t]
    \centering
    \captionsetup{font=footnotesize}
    \vspace{-0.5em}\includegraphics[width=1.0\linewidth]{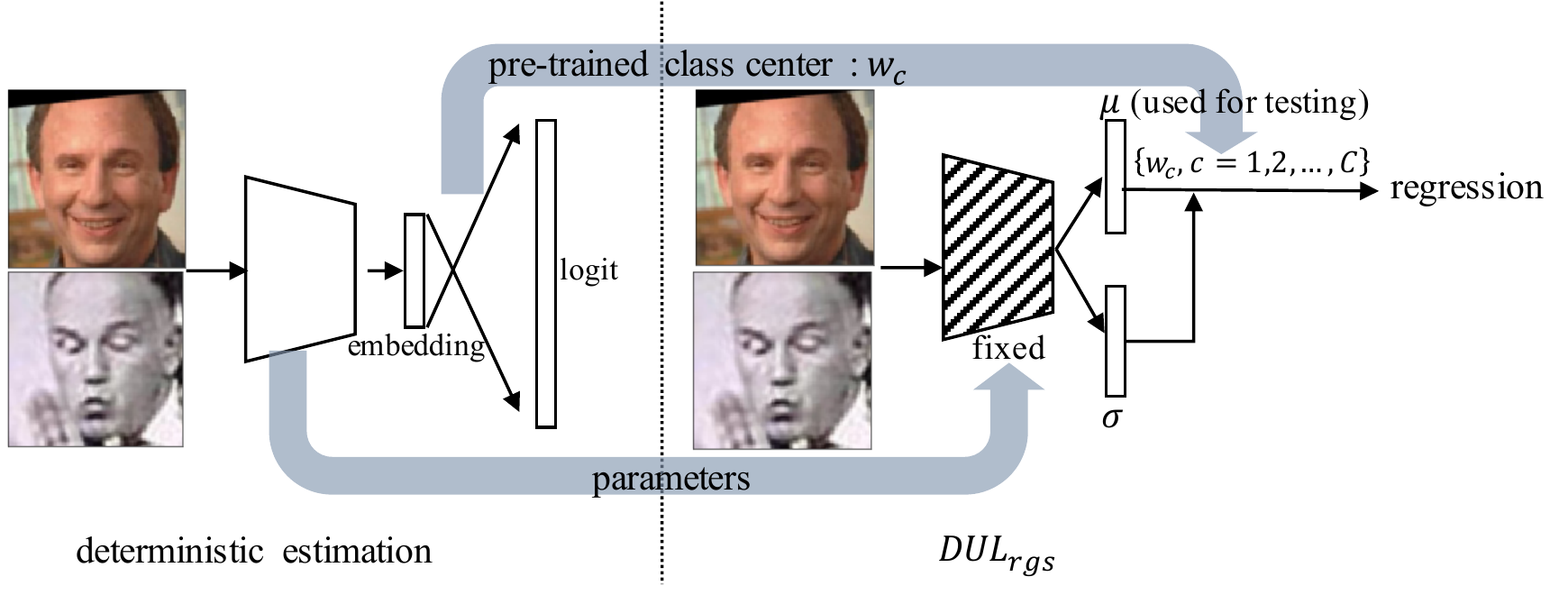}\\[-1.0em]
    \vspace{-0.2em}\caption{Overview of the proposed DUL\textsubscript{rgs} model. All parameters in the convolution layers are pre-trained by a deterministic FR model and are fixed during the training of DUL\textsubscript{rgs}.}\vspace{-1.0em}
    \label{fig:dul-rgs}
\end{figure}

\paragraph{Distributional Representation} Next we can estimate above $f(\bx_i)$ and $n(\bx_i)$ by data uncertainty regression. Specifically, a Gaussian distribution is assumed for the likelihood: $p(\bz_i|\bx_i) = \mathcal{N}(\bz_i; \bmu_i, \bsigma_i^2\mathbf{I})$ where $\bmu_i$ as well as $\bsigma_i$ are also parameterised by the weights in neural networks\footnote{Noted here $\bmu_i\approx f(\bx_i)$ and $\bsigma_i\approx n(\bx_i)$.} (see Fig.~\ref{fig:dul-rgs}). 
If we take each $\bw_c$ as the target, we should maximize the following likelihood for each $\bx_i$,
\begin{equation}
p(\bw_c | \bx_{i\in c}, \btheta)= \frac{1}{\sqrt{2\pi \boldsymbol{\sigma}_i^2}}\exp(-\frac{(\bw_c - \boldsymbol{\mu}_i)^2}{2\boldsymbol{\sigma}_i^2}).
\label{eq:likelihood}
\end{equation}
Actually, we take the log likelihood as follows,

\begin{equation}
\ln p(\bw_c | \bx_{i\in c}, \btheta) = - \frac{(\bw_c - \boldsymbol{\mu}_i)^2}{2\boldsymbol{\sigma}_i^2} - \frac{1}{2}\ln \boldsymbol{\sigma}_i^2 - \frac{1}{2}\ln 2\pi.
\label{eq:log-likelihood}
\end{equation}
Assumed that $\bx_i, i\in{1,2,...}$ are independently and identically distributed (\textit{iid.}), the likelihood over all data-points is $\prod_c\prod_i\ln p(\bw_c | \bx_{i\in c}, \btheta)$. Practically, we train the network to predict the log variance, $\br_i:=\ln \bsigma_i^2$, to stabilize the numerical during the stochastic optimization. Last, the likelihood maximization is reformulated as the minimization of cost function,
\begin{equation}
\mathcal{L}_{rgs} = \frac{1}{2N}\sum_{c}^{C}\sum_{i\in c}[\frac{1}{D}\sum_l^D( \exp(-r_i^{(l)})(w_c^{(l)} - \mu_i^{(l)})^2 + r_i^{(l)}],
\label{eq:loss-hum-rgs}
\end{equation}
where $D$, $N$ and $l$ refers to the size of embedding dimension, the size of data-points and the $l^{th}$ dimension of each feature vector, respectively. We omit the constant term, $\frac{D}{2}\ln 2\pi$ during the optimization.

\paragraph{Loss Attenuation Mechanism} By qualitatively analyzing Eq.~\ref{eq:log-likelihood}, our learned variance $\bsigma_i$ could actually be regarded as the uncertainty score measuring the confidence of the learned identity embedding, $\bmu_i$, belonging to $c^{th}$ class. 
Specifically, for those ambiguous $\bmu_i$ located far away from its class center $\bw_c$, DUL\textsubscript{rgs} will estimate large variance to temper the error term, $\frac{(\bw_c - \bmu)^2}{2\bsigma^2}$, instead of overfitting on these noisy samples. DUL\textsubscript{rgs} is discouraged from predicting large variance for all samples, which may lead to underfitting of $(\bw_c - \bmu)^2$ and larger $\log \bsigma$ term will punish the model in turn. Simultaneously, DUL\textsubscript{rgs} is also discouraged from predicting very small variance for all samples, which may lead to exponentially increases of error term. 
Thus, Eq.~\ref{eq:log-likelihood} allows DUL\textsubscript{rgs} to adapt the weighting of error term. This makes the model learn to attenuate the effect from those ambiguous $\bmu_i$ caused by poor quality samples.

\subsection{Discussion of Related Works}
\label{sec:discussion}
We first discuss the connection between DUL\textsubscript{cls} and variational information bottleneck (VIB)~\cite{alemi1612deep}.
VIB~\cite{alemi1612deep} is a variational approximation to information bottleneck (IB) principle~\cite{tishby2015deep} under the framework of deep learning. VIB seeks a stochastic mapping from input data $X$ to latent representation $Z$, in terms of the fundamental trade-off between making $Z$ as concise as possible but still have enough ability to predict label $Y$~\cite{tishby2015deep}. It is noted that $\mathcal{L}_{cls}$ is similar to the objective function in VIB. However, we analyze this classification method from data uncertainty perspective while VIB derived this objective function from the view of information bottleneck.

We next clarify some differences between DUL\textsubscript{rgs} and PFE~\cite{shi2019probabilistic}. Although both PFE and DUL\textsubscript{rgs} formally encode the input uncertainty as variance representation. However, PFE essentially measures the likelihood of each positive pair of $\{\bx_i, \bx_j\}$ sharing the same latent embedding: $p(\bz_i = \bz_j)$. While DUL\textsubscript{rgs} interprets a conventional \textit{Least-Square Regression} technique as a \textit{Maximum likelihood Estimation} with a data uncertainty regression model. 

Last, both DUL\textsubscript{cls} and DUL\textsubscript{rgs} 
\textbf{learn} identity representation $\bmu$ as well as uncertainty representation $\bsigma$, which ensure our predicted $\bmu$ can be directly evaluated by common-used matching metric. However, PFE has to use mutual likelihood score (\textit{MLS}) as the matching metric to improve the performance of deterministic model because identity representation is not learnt in PFE.
%------------------------------------------------------------------------
\section{Experiments}

In this section, we first evaluate the proposed methods on standard face recognition benchmarks. 
Then we provide qualitative and quantitative analysis to explore what is the meaning of the learned data uncertainty and how data uncertainty learning affects the learning of FR models. 
Last, we conduct experiments on the noisy MS-Celeb-1M dataset to demonstrate that our methods perform more robustly than deterministic methods.

\begin{table*}[t]
\captionsetup{font=footnotesize}
\newcommand{\mr}[1]{\multirow{2}{*}{#1}}
\footnotesize
\setlength{\tabcolsep}{9pt}
\begin{center}
\begin{tabularx}{\linewidth}{Xlcccc cccc}
\toprule
\mr{Base Model} & \mr{Representation} & \mr{LFW}  & \mr{CFP-FP} & \mr{MegaFace(R1)}   & \mr{YTF} & \multicolumn{4}{c}{IJB-C (TPR@FPR)} \\
\cline{7-10}\\[-1.0em]
                &                     &           &             &       &          & $0.001\%$   & $0.01\%$  & $0.1\%$ & AUC\\
\midrule

\rowcolor{Gray}
& Original  & $99.63$  & $96.85$ & $97.11$ & $96.09$ & $75.43$ & $88.65$  & $94.73$ & $87.51$\\
\rowcolor{Gray}
& PFE~\cite{shi2019probabilistic}  & $99.68$ & $94.57$ & $97.18$ & $96.12$ & $86.24$ & $92.11$  & $\bold{95.71}$ & $91.71$\\
\rowcolor{Gray}
\multirow{-2}{*}{AM-Softmax~\cite{wang2018additive}}  
& DUL\textsubscript{cls}  & $\bold{99.71}$  & $97.28$ & $\bold{97.30}$ & $\bold{96.46}$  & $\bold{88.25}$ & $\bold{92.78}$ & $95.57$ & $\bold{92.40}$\\
\rowcolor{Gray}
& DUL\textsubscript{rgs}       & $99.66$  & $\bold{97.61}$ & $96.85$ & $96.28$ & $87.02$ & $91.84$ & $95.02$ & $91.44$ \\

& Original  & $99.64$  & $96.77$ & $97.08$ & $96.06$ & $73.80$ & $88.78$ & $94.30$ & $86.94$\\
& PFE~\cite{shi2019probabilistic}  & $99.68$  & $95.34$ & $96.55$ & $96.32$ & $86.69$ & $92.28$  & $\bold{95.66}$ & $91.89$\\
\multirow{-2}{*}{ArcFace~\cite{deng2019arcface}}
& DUL\textsubscript{cls}   & $\bold{99.76}$  & $97.01$ & $\bold{97.22}$ & $96.20$ & $\bold{87.22}$ & $\bold{92.43}$ & $95.38$ & $\bold{92.10}$\\
& DUL\textsubscript{rgs}      & $99.66$ & $\bold{97.11}$ & $96.83$ & $\bold{96.38}$  & $86.21$ & $91.03$ & $94.53$ & $90.79$ \\

\rowcolor{Gray}
& Original  & $99.60$ & $95.87$ & $90.34$ & $95.89$ & $77.60$ & $86.19$ & $92.55$ & $85.83$\\
\rowcolor{Gray}
& PFE~\cite{shi2019probabilistic}  & $99.66$  & $86.45$ & $90.64$ & $95.98$  & $79.33$ & $87.28$  & $93.41$ & $87.01$\\
\rowcolor{Gray}
\multirow{-2}{*}{L2-Softmax~\cite{ranjan2017l2}}  
& DUL\textsubscript{cls} & $99.63$ & $\bold{97.24}$ & $\bold{93.19}$ & $\bold{96.56}$ & $\bold{79.90}$ & $\bold{87.80}$ & $\bold{93.44}$ & $\bold{87.38}$\\
\rowcolor{Gray}
& DUL\textsubscript{rgs}   & $99.66$ & $96.35$ & $89.66$ & $96.08$ & $74.46$ & $83.23$ & $91.09$ & $83.10$\\

\bottomrule
\end{tabularx}
\vspace{-0.9em}\caption{Results of models (ResNet18) trained on MS-Celeb-1M. ``Original'' refers to the deterministic embeddings. The better performance among each base model are shown in bold numbers. We use $\bsigma$ both for fusion and matching (with mutual likelihood scores) in PFE. AUC is calculated when FPR spans on the interval [$0.001\%$, $0.1\%$] and we rescale it.}
\vspace{-2.5em}
\label{tab:compare-1}
\end{center}
\end{table*}
%-------------------------------------------------------------------------
\subsection{Datasets and Implementation Details}
\label{sec: implementation}
We describe the public datasets that are used, and our implementation details.

\paragraph{Datasets} We use MS-Celeb-1M datasets with 3,648,176 images of 79,891 subjects as training set. 2 benchmarks including LFW~\cite{huang2008labeled} and MegaFace~\cite{kemelmacher2016megaface}\footnote{Noted that we use rank1 protocol of MegaFace}, and 3 unconstrained benchmarks: CFP~\cite{sengupta2016frontal}\footnote{Noted that we only use ``frontal-profile'' protocol of CFP}, YTF~\cite{wolf2011face} and IJB-C~\cite{maze2018iarpa}, are used to evaluate the performance of DUL\textsubscript{cls/rgs} following the standard evaluation protocols. 
\paragraph{Architecture} We train baseline models on ResNet~\cite{he2016deep} backbone with SE-blocks~\cite{hu2018squeeze}. The head of the baseline model is: \texttt{BackBone-Flatten-FC-BN} with embedding dimensions of $512$ and dropout probability of $0.4$ to output the embedding feature. Compared with baseline model, DUL\textsubscript{cls} has an additional head branch sharing the same architecture to output the variance. DUL\textsubscript{rgs} also has an additional head branch whilst its architecture is:  \texttt{BackBone-Flatten-FC-BN-ReLU-FC-BN-exp}, to output the variance.

\paragraph{Training} All baseline models and DUL\textsubscript{cls} models are trained for 210,000 steps using a SGD optimizer with a momentum of $0.9$, weight decay of $0.0001$, batch size of $512$. We use triangular learning rate policy~\cite{smith2017cyclical} with the $max\_lr$ of $0.1$ and $base\_lr$ of $0$. For most DUL\textsubscript{cls} models, we set trade-off hyper-parameter $\lambda$ as $0.01$. 
For the proposed DUL\textsubscript{rgs}, we first train baseline model for 210,000 steps and then fix parameters in all convolution layers (\textit{step} 1). Then we train the mean branch as well as the variance branch in head from scratch for additional 140,000 steps with batch size of $256$ (\textit{step} 2). During \textit{step 2}, we set learning rate starting at $0.01$, and then decreased to $0.001$ and $0.0001$ at 56,000 and 84,000 steps.

%-------------------------------------------------------------------------
\subsection{Comparing DUL with Deterministic Baselines}
\label{sec:baseline}
In this part, all baseline models are trained with ResNet18 backbone~\cite{he2016deep}, equipped with different variants of softmax loss, i.e., AM-Softmax~\cite{wang2018additive}, ArcFace~\cite{deng2019arcface} and L2-Softmax~\cite{ranjan2017l2}. Both the embedding features and the weights in classifier are $\ell$2-normalized during the training. Our proposed DUL\textsubscript{cls} models are trained with the same backbone and loss functions. Our proposed DUL\textsubscript{rgs} models are trained based on the different pre-trained baseline models, as described in Section~\ref{sec: implementation}.

Table~\ref{tab:compare-1} reports the testing results obtained by the baseline models (``Original'') and the proposed DUL models. Cosine similarity is used for evaluation. 
Our proposed methods outperform the baseline deterministic models on most benchmarks\footnote{Noted that DUL\textsubscript{rgs} combined with L2-Softmax deteriorates on IJB-C, which should be further explored in the future.}. This demonstrates that the proposed methods are effective on different state-of-the-art loss functions. These results indicate that the identity embeddings ($\bmu$ in our methods) trained with data uncertainty ($\bsigma$ in our method) present better intra-class compactness and inter-class separability than the point embeddings estimated by baseline models, especially on those unconstrained benchmarks: CFP with frontal/profile photos and YTF/IJB-C with most blur photos collected from YouTube videos, compared with benchmarks with most clear and frontal photos (LFW and MegaFace).

%------------------------------------------------------------------------
\begin{figure}[t]
    \centering
    \captionsetup{font=footnotesize}
    \vspace{-0.5em}
    \includegraphics[width=1.0\linewidth]{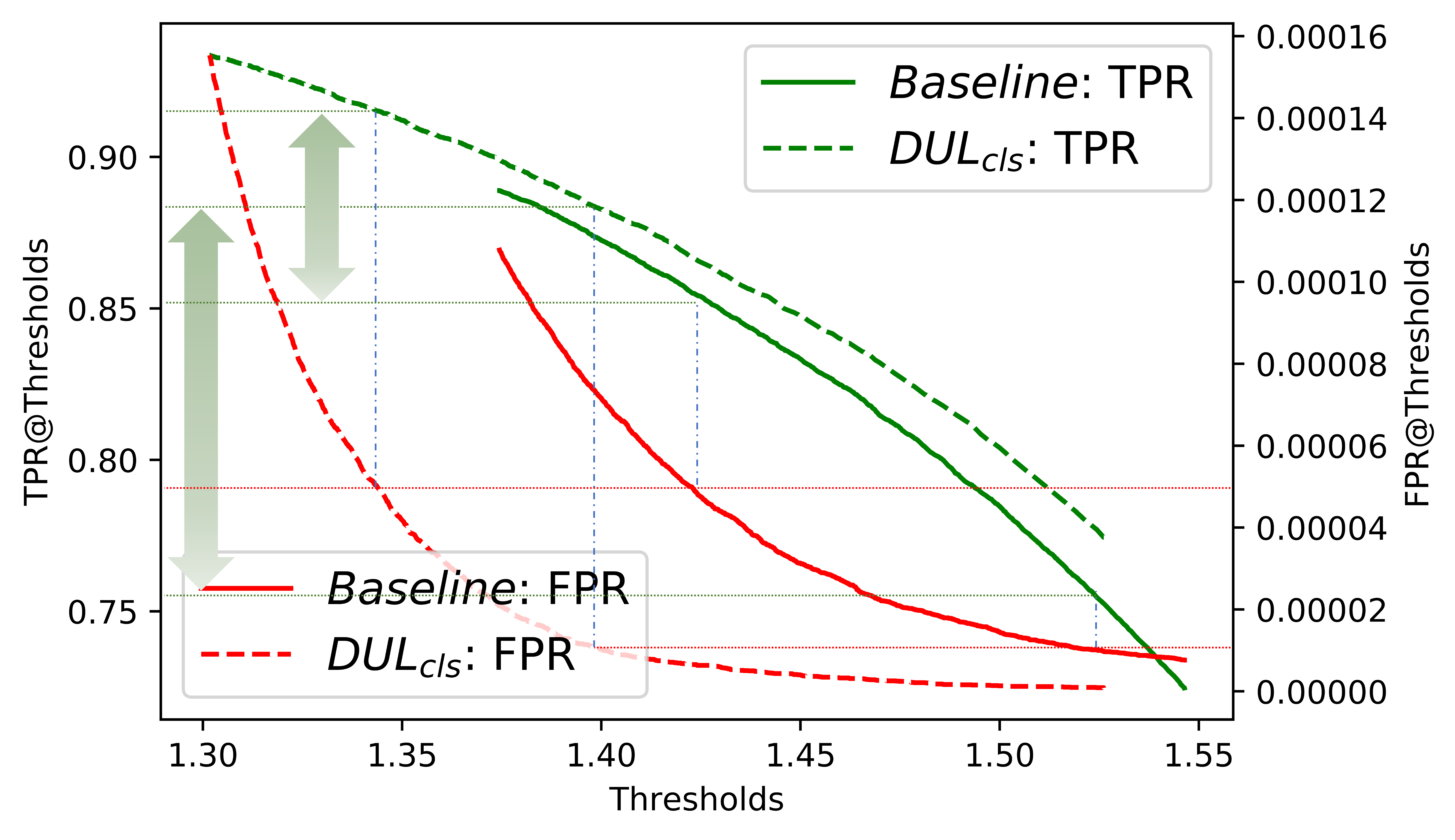}\\[-0.0em]
    \includegraphics[width=1.0\linewidth]{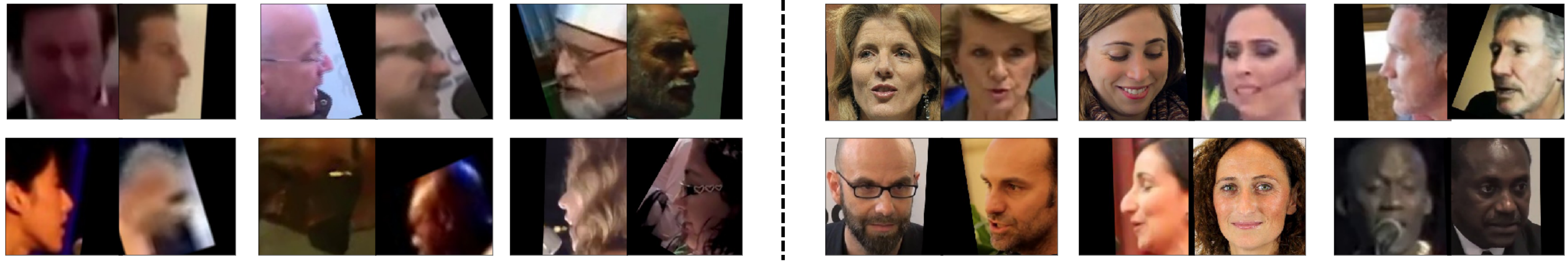}\\[-1.0em]
    \vspace{0.3em}
    \caption{Top: TPR\&FPR \textit{vs.} threshold on IJB-C; Bottom: false acceptance cases mostly happened in the baseline model (left); false acceptance cases mostly happened in DUL\textsubscript{cls} (right). Both baseline model and DUL\textsubscript{cls} are trained by ResNet18 with AM-Softmax on MS-Celeb-1M dataset. \textbf{Best viewed in color}.}
    \vspace{-1.0em}
    \label{fig:tpr@fpr}
\end{figure}

The proposed DUL achieves most remarkable improvement on verification protocols of IJB-C benchmark, which is also the most challenging one. We thus plot how true acceptance rate (TPR) and false acceptance rate (TPR) perform along with the change of thresholds.
As illustrated in Fig~\ref{fig:tpr@fpr}, DUL\textsubscript{cls} achieves higher TPR and lower FPR than baseline model at different settings of matching threshold. Additionally, the lower FPR is set, the better DUL\textsubscript{cls} performs on TPR. 
Fig~\ref{fig:tpr@fpr} also shows the vast majority cases of false acceptance respectively happened in baseline model and DUL\textsubscript{cls}. We can see that DUL\textsubscript{cls} resolves more FP cases with extreme noises, which are typically occurring in the baseline model.
This indicates that model with data uncertainty learning is more applicable to the unconstrained face recognition scenario than deterministic model. 

We have the similar conclusion for DUL\textsubscript{rgs}. 

\subsection{Comparing DUL with PFE}

For comparison, we re-implemented PFE on all baseline models according to the recommended settings of implementation details in ~\cite{shi2019probabilistic}\footnote{which means we use mutual likelihood score (\textit{MLS}) for matching and its proposed fusion strategy for feature aggregation in template/video benchmarks, i.e., YTF and IJB-C.}. We note that our re-implementation has achieved similar or slightly better results than those in~\cite{shi2019probabilistic}. Our DUL\textsubscript{cls/rgs} use averaged pooling aggregation for features in template and are evaluated by cosine similarity.
Compared with PFE, our proposed DUL\textsubscript{cls} achieves better performances in all cases, and the proposed DUL\textsubscript{rgs} also shows competitive performances. Results are reported in Table~\ref{tab:compare-1}.

PFE interprets the point embedding learned by deterministic FR models as the mean of its output distributional estimation and only learn the uncertainty (variance) for each sample. Thus, PFE has to use \textit{MLS} metric, which takes the predicted variance into account. Although PFE achieves better results with the help of the matching measurement with more precision, it still suffers more computational complexity for matching. Specifically, for verification of 6000 face pairs (LFW), standard cosine metric takes less than 1 second via matrix multiplication, while \textit{MLS} takes 1min28s, on two GTX-1080.

% ------------------------------------------------------------------------
\subsection{Comparison with State-Of-The-Art}
To compare with state-of-the-art, we use a deeper and stronger backbone, ResNet64, trained with AM-Softmax loss on MS-Celeb-1M dataset, as our baseline model. Then we train the proposed DUL models following the setting described in section~\ref{sec: implementation}. 

The results are illustrated in Table~\ref{tab:lfw}. Noted that performances of baseline model have been saturated on LFW and CFP-FP, where the merit of data uncertainty learning is not obvious. However, DUL\textsubscript{cls/rgs} still slightly improve the accuracy on YTF and MegaFace\footnote{Noted that our used MegaFace datasets is refined, while previous reported SOTA results in Table~\ref{tab:lfw} usually use non-refined MegaFace.}.
Table~\ref{tab:ijbc} reports the results of different methods on IJB-C. Both PFE and DUL achieve much better performances over baseline models.

\begin{table}[t]
\captionsetup{font=footnotesize}
\newcommand{\mr}[1]{\multirow{2}{*}{#1}}
\setlength{\tabcolsep}{4pt}
\footnotesize
\begin{center}
\resizebox{1.0\columnwidth}{!}{%
\begin{tabularx}{1.0\linewidth}{Xccccc}
\toprule
Method      & Training Data     & LFW & YTF & MegaFace   & CFP-FP \\
\midrule            
FaceNet~\cite{schroff2015facenet}       & 200M  & $99.63$   & $95.1$    & -  & -        \\
DeepID2+~\cite{sun2015deeply}             & 300K  & $99.47$   & $93.2$    & -  & -        \\
CenterFace~\cite{wen2016discriminative} & 0.7M  & $99.28$   & $94.9$    & $65.23$ & $76.52$    \\
SphereFace~\cite{liu2017sphereface}    & 0.5M  & $99.42$   & $95.0$    & $75.77$ & $89.14$   \\
ArcFace~\cite{deng2019arcface}        & 5.8M  & $99.83$  & $\bold{98.02}$    & $81.03$   & $96.98$    \\
CosFace~\cite{wang2018cosface}         & 5M    & $99.73$   & $97.6$    & $77.11$ & $89.88$    \\
L2-Face~\cite{ranjan2017l2}           & 3.7M  & $99.78$   & $96.08$   & -  & -        \\
Yin~\etal~\cite{yin2017multi} & 1M  & $98.27$   & -   & -  & $94.39$       \\
PFE~\cite{shi2019probabilistic} & 4.4M  & $99.82$   & $97.36$   & $78.95$  & $93.34$       \\
\hline
Baseline                                & 3.6M  & $99.83$   & $96.50$   & $98.30$   & $98.75$   \\
PFE\textsubscript{rep}          & 3.6M  & $99.82$ & $96.50$   & $98.48$ & $97.28$    \\
DUL\textsubscript{cls}                 & 3.6M  & $99.78$    & $96.78$   & $98.60$  & $98.67$      \\
DUL\textsubscript{rgs}           & 3.6M  & $99.83$ & $96.84$  & $98.12$ & $98.78$    \\
% DUL\textsubscript{rgs+mls}           & 3.6M  & $99.81$ & $96.80$  & $98.15$ & $98.12$    \\
\bottomrule
\end{tabularx}
}
\vspace{-1.2em}\caption{Comparison with the state-of-the-art methods on LFW, YTF, MegaFace (MF) and CFP-FP. ``-'' indicates that the author did report the performance on the corresponding protocol. ``PFE\textsubscript{rep}'' means we reproduce PFE by ourself. Backbone: ResNet64.}\vspace{-2.8em}
\label{tab:lfw}
\end{center}
\end{table}

%------------------------------------------------------------------------

\begin{table}[t]
\newcommand{\hly}{\cellcolor{Y}}
\newcommand{\hlg}{\cellcolor{G}}
\captionsetup{font=footnotesize}
\newcommand{\mr}[1]{\multirow{2}{*}{#1}}
\footnotesize
\setlength{\tabcolsep}{4.6pt}
\begin{center}
\begin{tabularx}{1.00\linewidth}{Xc cccc}
\toprule
\mr{Method}                                  & \mr{Training Data} & \multicolumn{4}{c}{IJB-C (TPR@FPR)} \\
\cline{3-6}\\[-1.0em]
 &           & $0.001\%$   & $0.01\%$  & $0.1\%$ & AUC\\
\midrule            
Yin~\etal~\cite{yin2018towards}        & 0.5M      & - & - & $69.3$ & - \\
Cao~\etal~\cite{cao2018vggface2}       & 3.3M      & $74.7$ & $84.0$ & $91.0$ & - \\
Multicolumn~\cite{xie2018multicolumn}   & 3.3M      & $77.1$ & $86.2$ & $92.7$ & - \\
DCN~\cite{xie2018comparator}           & 3.3M      & - & $88.5$ & $94.7$ & - \\
PFE~\cite{shi2019probabilistic}               & 4.4M      & $89.64$ & $93.25$   & $95.49$ & -\\
\hline
Baseline                                & 3.6M      & $83.06$ & $92.16$   & $95.83$ & $91.97$\\
PFE\textsubscript{rep}           & 3.6M      & $89.77$ & $94.14$   & $96.37$ & $93.74$ \\
DUL\textsubscript{cls}               & 3.6M      & $88.18$ & $\bold{94.61}$   & $\bold{96.70}$ & $\bold{93.97}$\\
DUL\textsubscript{rgs}          & 3.6M      & $\bold{90.23}$ & $94.21$   & $96.32$ & $93.88$ \\
% DUL\textsubscript{rgs+mls}          & 3.6M      & $\bold{92.46}$ & $\bold{95.19}$   & $\bold{96.98}$ & $\bold{95.07}$ \\
\bottomrule
\end{tabularx}
\vspace{-1.2em}\caption{Comparison with the state-of-the-art methods on IJB-C. Backbone: ResNet64. }\vspace{-2.8em}
\label{tab:ijbc}
\end{center}
\end{table}

\subsection{Understand Uncertainty Learning}
In this part, we qualitatively and quantitatively analyze the proposed DUL to gain more insights about data uncertain learning. 

\begin{figure}[t]
    \centering
    \captionsetup{font=footnotesize}
    \vspace{0.5em}
    \includegraphics[width=0.8\linewidth]{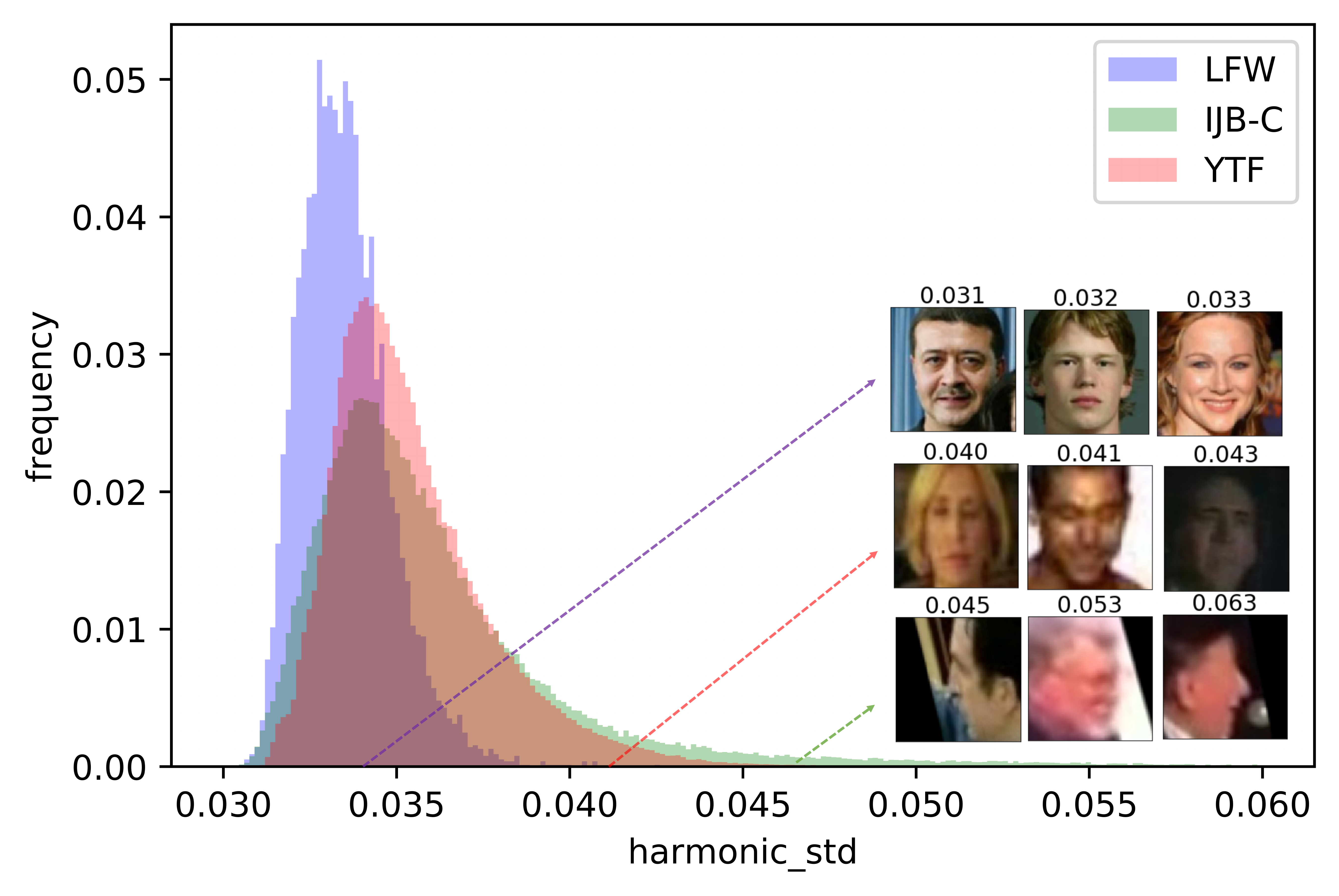}\\[-1.0em]
    \vspace{-0.2em}\caption{ Uncertainty distribution on different dataset for DUL\textsubscript{rgs}. Similar uncertainty distribution has also been observed in DUL\textsubscript{cls}. \textbf{Best viewed in color}.}
    \vspace{-1.0em}
    \label{fig:uncertainty-rgs}
\end{figure}

\paragraph{What is the meaning of the learned uncertainty?}
The estimated uncertainty is closely related to the quality of face images, for both DUL\textsubscript{cls} and DUL\textsubscript{rgs}. This is also observed in PFE~\cite{shi2019probabilistic}. For visualization, we show the learned uncertainty\footnote{Specifically, we use harmonic mean of the predicted variance $\bsigma\in\mathbb{R}^{512}$ as the approximated measurement of the estimated uncertainty. The same below.} of different dataset in Figure~\ref{fig:uncertainty-rgs}. It illustrates that the learned uncertainty increases along with the image quality degradation. This learned uncertainty could be regarded as the quality of the corresponding identity embedding estimated by the model, measuring the proximity of the predicted face representation to its genuine (or true) point location in the latent space.

Therefore, two advantages are obtained for face recognition with data uncertainty learning. First, the learned variance can be utilized as a ``risk indicator'' to alert FR systems that the output decision is unreliable when the estimated variance is very high. Second, the learned variance also can be used as the measurement of image quality assessment. In this case, we note that it is unnecessary to train a separate quality assessment model which requires explicit quality labels as before.

%------------------------------------------------------------------------

\paragraph{How the learned uncertainty affect the FR model?} 
In this part, we attempt to shed some light on the mechanism of how the learned data uncertainty affects the model training and helps to obtain better feature embeddings.

\begin{figure}[t]
    \centering
    \captionsetup{font=footnotesize}
    \vspace{-0.5em}\includegraphics[width=0.8\linewidth]{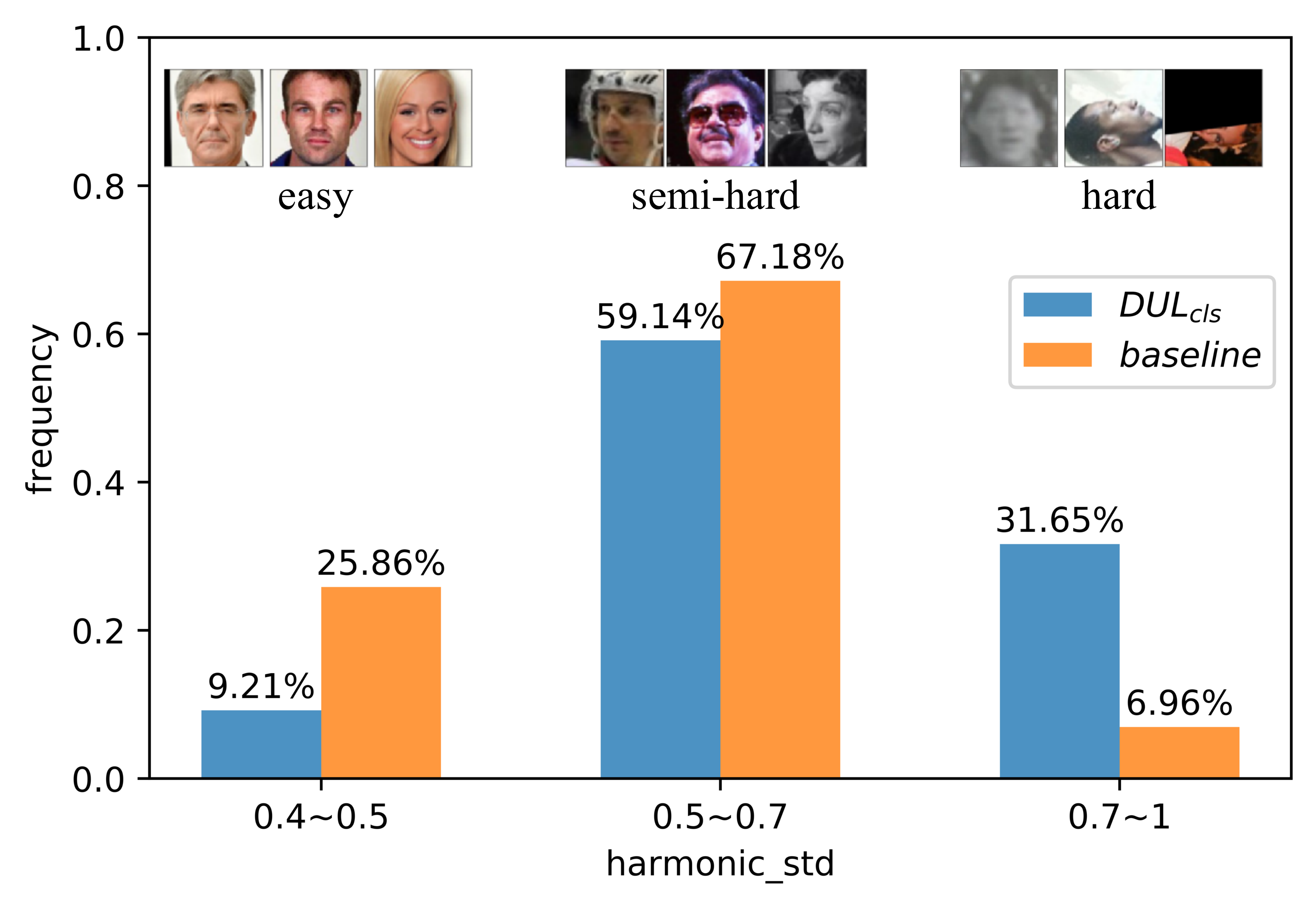}\\[-1.0em]
    \vspace{-0.2em}\caption{Bad case analysis between baseline model and DUL\textsubscript{cls}. \textbf{Best viewed in color}.}
    \vspace{-1.0em}
    \label{fig:bad-case}
\end{figure}

We classify the training samples in MS-Celeb-1M dataset into three categories according to the degree of estimated uncertainty by DUL\textsubscript{cls}: \emph{easy} samples with low variance, \emph{semi-hard} samples with medium variance and \emph{hard} samples with large variance.
We calculated the proportion of mis-classified samples in each of the three categories to all mis-classified samples respectively produced by baseline model and our DUL\textsubscript{cls}. 
Fig~\ref{fig:bad-case} illustrates that our DUL\textsubscript{cls} causes relatively less bad cases on easy samples as well as semi-hard samples, compared with the basline model. However, for those hard samples with extreme noises, baseline model produces less bad cases, when compared with DUL\textsubscript{cls}. This demonstrates that FR networks with data uncertainty learning focus more on those training samples which \textbf{should} be correctly classified and simultaneously ``give up'' those detrimental samples, instead of over-fitting them. This supports our previous discussion in Section~\ref{sec:classification}.

\begin{figure}[t]
    \centering
    \captionsetup{font=footnotesize}
    \vspace{-0.5em}\includegraphics[width=1.0\linewidth]{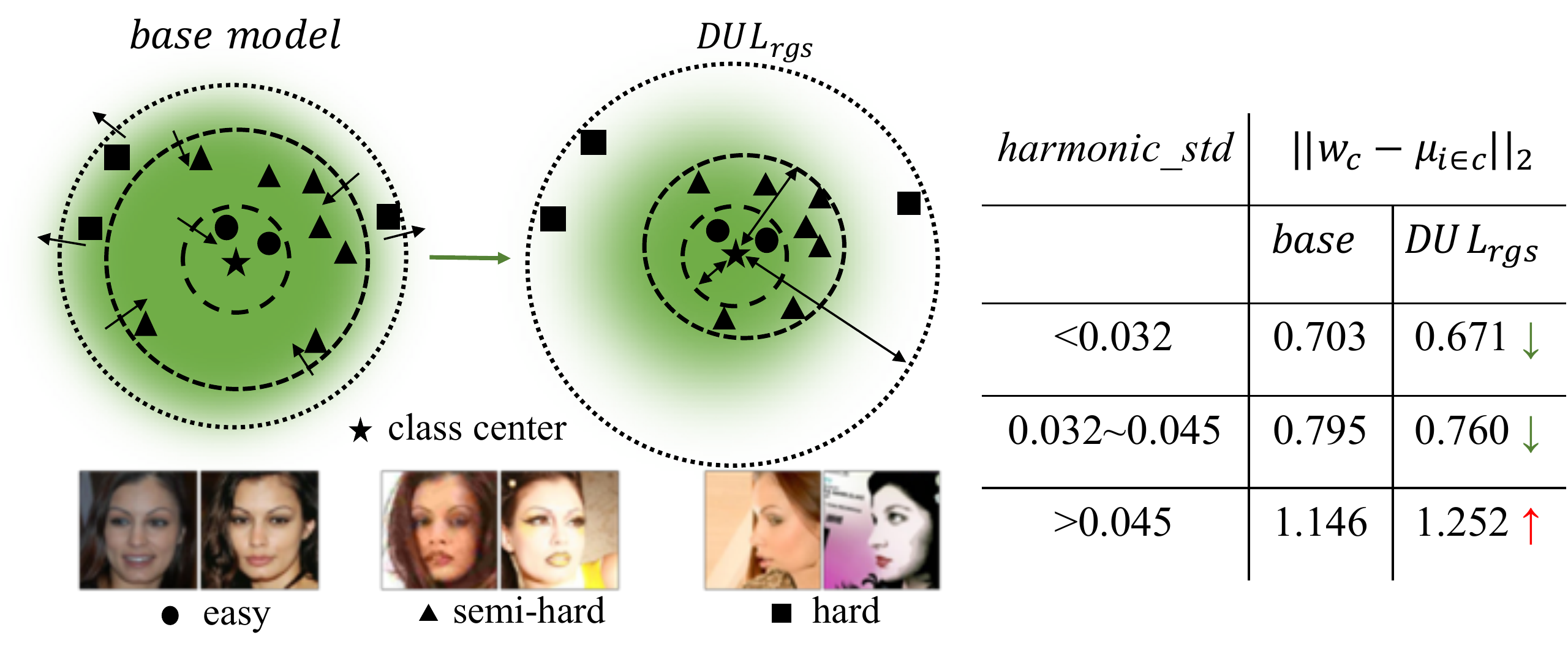}\\[-1.0em]
    \vspace{-0.1em}\caption{Averaged intra-class distances $||\bmu_{i\in c}-\bw_c||_2$ between base model and DUL\textsubscript{rgs}.}
    \vspace{-1.0em}
    \label{fig:anti-overfitting}
\end{figure}

We also conduct similar experiment for DUL\textsubscript{rgs}. We calculate the averaged euclidean distances\footnote{Noted this averaging distances are further averaged over all classes in MS-Celeb-1M.} between the class center $\bw_c$ and its intra-class estimated identity embedding, $\bmu_{i\in c}$, respectively for baseline model and DUL\textsubscript{rgs}. As illustrated in Fig~\ref{fig:anti-overfitting}, DUL\textsubscript{rgs} pulls the easy and semi-hard samples closer to their class center whilst pushes those hard samples further away. This also supports our discussion in Section~\ref{sec:regression} that Eq.~\ref{eq:log-likelihood} effectively prevents model over-fitting on extremely noisy samples by the adaptive weighting mechanism w.r.t $\bsigma$.

\begin{figure}[t]
    \centering
    \captionsetup{font=footnotesize}
    \begin{minipage}{0.35\linewidth}
    \includegraphics[width=1.0\linewidth]{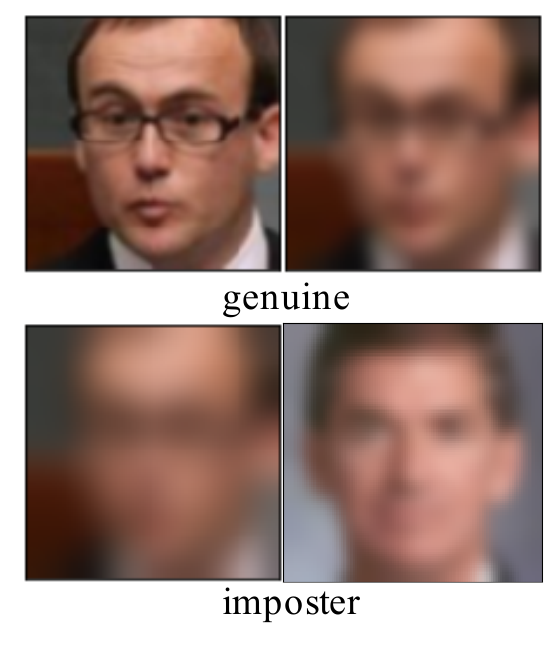}\\[-2.0em]
    \begin{center} \footnotesize(a)\end{center}\vspace{-1.2em}
    \end{minipage}\hfill
    \begin{minipage}{0.65\linewidth}
    \includegraphics[width=1.0\linewidth]{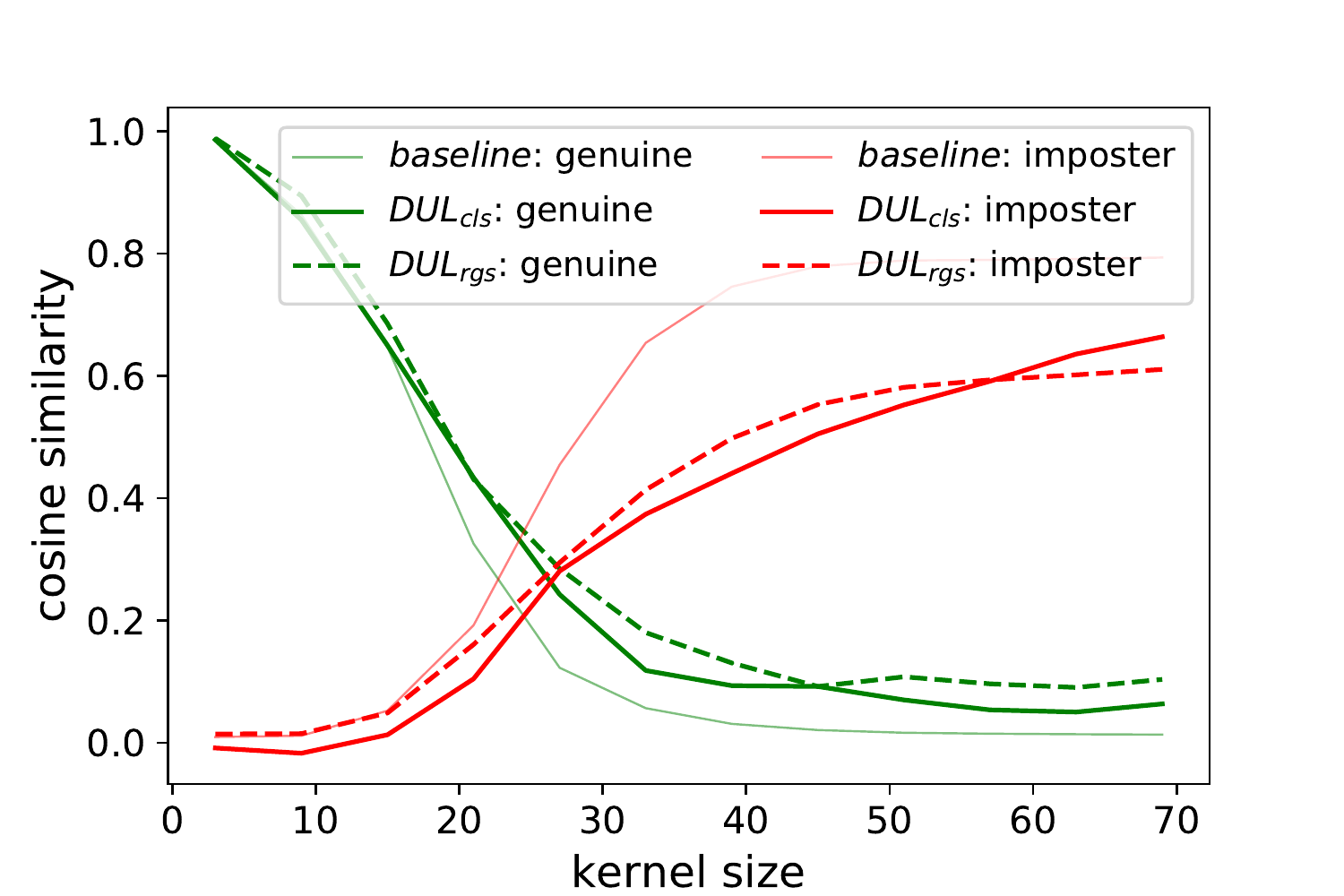}\\[-2.0em]
    \begin{center} \footnotesize(b) \end{center}\vspace{-1.2em}
    \end{minipage}\hfill
    \vspace{-0.3em}\caption{(a) Blur genuine and imposter pair; (b) Cosine similarity score obtained by baseline and proposed DUL for each pair.}\vspace{-1.0em}
    \label{fig:dilemma}
\end{figure}

Last, we manually construct imposter/genuine test pair with different blurriness to compare the cosine similarity respectively obtained by baseline model and our methods. As illustrated in Fig~\ref{fig:dilemma}, along with the increase of blurriness, both baseline model and DUL deteriorate rapidly. However, our proposed DUL achieves higher similarity score for genuine pair and lower similarity score for imposter pair than baseline model, indicating that it is more robust.

%------------------------------------------------------------------------

\subsection{Other Experiments}
\label{sec: other-experiment}

\paragraph{Impact of hyper-parameter of DUL\textsubscript{cls}} 
In this part, we qualitatively analyze what the trade-off hyper-parameter $\lambda$ controls in DUL\textsubscript{cls}. 
As mentioned in VIB~\cite{alemi1612deep}, KLD term works as a regularization to trade off the conciseness and the richness of the information preserved in bottleneck embeddings. We experimentally find the KL divergence in our method affects the representational capacity of $\bsigma$. 
As illustrated in Table~\ref{tab:trade-off-lambda}, DUL\textsubscript{cls} without the optimization of KLD term ($\lambda=0$) performs close to baseline model. In this case, DUL\textsubscript{cls} estimates relatively small $\bsigma_i$ for all samples, which makes the sampled representation $\bmu_i+\epsilon\bsigma_i$ nearly deterministic. With the enhancement of the optimization strength of KLD term ($\lambda\uparrow$), DUL\textsubscript{cls} is prone to ``assign'' larger variance for noisy samples and small variance for high quality ones (as illustrated in Fig~\ref{fig:bad-case}). However, overly minimizing KLD ($\lambda=1$) term will 
prompt the model to predict large variance for all samples, which makes $\mathcal{L}_{cls}$ in Eq.~\ref{eq:softmax-loss} hard to converge, thus the performances deteriorate rapidly (see Table~\ref{tab:trade-off-lambda}).

\begin{table}[t]
\newcommand{\hly}{\cellcolor{Y}}
\newcommand{\hlg}{\cellcolor{G}}
\captionsetup{font=footnotesize}
\newcommand{\mr}[1]{\multirow{2}{*}{#1}}
\footnotesize
\setlength{\tabcolsep}{4.6pt}
\begin{center}
\begin{tabularx}{1.00\linewidth}{Xccc ccc}
\toprule
\mr{$\lambda$} &\mr{$\overline{\bsigma}$}   & \mr{YTF} &\mr{MegaFace} & \multicolumn{3}{c}{IJB-C (TPR@FPR)} \\
\cline{5-7}\\[-1.0em]
 &        &  &  & $0.001\%$   & $0.01\%$  & $0.1\%$ \\
\midrule            
baseline  &  -  &  $96.09$           & $97.11$           & $75.32$          & $88.65$  & $94.73$ \\
\hline
$0.0$   & $0.2562$  &  \better{$96.14$}  & \better{$97.13$}  & \worse{$64.92$}  & \worse{$88.55$}   & \worse{$94.64$}  \\
$0.0001$    & $0.3074$  &  \better{$96.36$}  & \better{$97.25$}  & \worse{$65.44$}  & \worse{$85.22$}   & \worse{$94.44$} \\
$0.001$     & $0.3567$  &  \better{$96.26$}  & \better{$97.38$}  & \worse{$62.88$}  & \worse{$86.65$}   & \worse{$94.46$} \\
$0.01$      & $0.5171$  &  \better{$96.46$}  & \better{$97.30$}  & \better{$88.25$} & \better{$92.78$}  & \better{$95.57$}  \\
$0.1$       & $0.8505$  &  \better{$96.42$}  & \worse{$95.07$}  & \better{$87.19$}  & \better{$91.78$}  & \better{$95.13$} \\
$0.5$       & $0.9012$  &  \worse{$87.40$}  & \worse{$85.73$}  & \worst{$40.23$}  & \worst{$52.70$}  & \worst{$58.52$} \\
$1.0$       & $0.9520$  &  \worst{$75.14$}   & \worst{$63.90$}  & \worst{$1.770$}   & \worst{$4.530$}     & \worst{$13.02$} \\
\bottomrule
\end{tabularx}
\vspace{-0.7em}\caption{Results of DUL\textsubscript{cls} trained with different trade-off $\lambda$. $\overline{\bsigma}$ represents we average the harmonic mean of the estimated variance over all training samples in MS-Celeb-1M. The backbone is ResNet18 with AM-Softmax loss.}
\vspace{-2.8em}
\label{tab:trade-off-lambda}
\end{center}
\end{table}

%------------------------------------------------------------------------

\begin{table}[]
\captionsetup{font=footnotesize}
%\tiny
\footnotesize
\setlength{\tabcolsep}{2.8pt}
\begin{tabularx}{1.00\linewidth}{ccc ccccc}
\toprule
\multirow{2}{*}{ percent } & \multirow{2}{*}{Model} & \multirow{2}{*}{MegaFace} & \multirow{2}{*}{LFW} & \multirow{2}{*}{YTF}& \multicolumn{3}{c}{IJB-C (TPR@FPR)} \\ \cline{6-8} \\[-1.0em]
                  &                   &                   &                   &                   &  $0.001\%$    &     $0.01\%$    &      $0.1\%$   \\
\hline
0\% &         baseline          &     97.11              &     99.63 & 96.09    &    75.32   &    88.65   &  94.73     \\ \cline{1-8}
\hline
\multirow{4}{*}{10\%} &    baseline       &    96.64     &    99.63  &96.16  &   64.96    &    86.00   &   94.82    \\ 
                                 &   PFE~\cite{shi2019probabilistic}    &   97.02     &     99.63  &96.1 &    83.39  &    91.33   &    95.54  \\ 
                                     &   DUL\textsubscript{cls}  &     96.88     &    99.75   &96.44  &     88.04  &     93.21  &     95.96   \\
                                     &    DUL\textsubscript{rgs}  &     96.05     &    99.71  &96.46   &     84.74  &     91.56  &     95.30  \\ \cline{1-8}
\hline
                  \multirow{4}{*}{20\%} &     baseline              &    96.20       &   99.61  &96.00 &    43.52   &    80.48   &    94.22   \\
                                        &   PFE~\cite{shi2019probabilistic}    &   96.90     &     99.61  &95.86 &     82.03  &    90.89   &     95.38  \\ 
                                     &   DUL\textsubscript{cls}    &   96.37     &     99.71  &96.68 &     89.01  &    93.24   &     95.97  \\ 
                                      &   DUL\textsubscript{rgs}    &   95.51     &     99.66  &96.64 &     81.10  &    90.91   &     95.27  \\  \cline{1-8}
\hline
                   \multirow{4}{*}{30\%} &     baseline                &  95.72          &   99.60  &95.45 &    31.51   &   76.09    &    93.11  \\ 
                                 &   PFE~\cite{shi2019probabilistic}    &   96.82     &     99.61  &96.12 &    80.92  &    90.31   &    95.29  \\ 
                                     &   DUL\textsubscript{cls}   &      95.86    &    99.73   &96.38 &    86.05   &    91.80   &    95.02   \\ 
                                      &   DUL\textsubscript{rgs}    &   94.96     &     99.66  &96.66 &     81.54  &    91.20   &     95.32  \\  \cline{1-8}
\hline
                  \multirow{4}{*}{40\%} &     baseline       &   95.14       &     99.56 &95.51  &  39.69     &  77.12     &  93.73     \\ 
                                  &   PFE~\cite{shi2019probabilistic}    &   96.59     &     99.59  &95.94 &    77.72  &    89.46   &    94.82  \\ 
                                     &  DUL\textsubscript{cls}   &     95.33  &   99.66 &96.54  &   84.15    &   92.60    &   95.85    \\ 
                                      &   DUL\textsubscript{rgs}    &   94.28     &     99.58   &96.68 &     78.13  &    87.64   &     94.67  \\ 
\bottomrule
\end{tabularx}
\vspace{-0.7em}\caption{Comparison of baseline model and proposed DUL\textsubscript{cls/rgs} trained on noisy MS-Celeb-1M. Backbone model is ResNet18 with AM-Softmax loss.}
\vspace{-2.0em}
\label{tab:noise-trainset}
\end{table}

\paragraph{DUL performs more robustly on noisy training data.} 
Based on the analysis of Section~\ref{sec:discussion} about how the learned variance affect the model training. We further conduct experiments on noisy MS-Celeb-1M to prove it. We randomly select different proportions of samples from MS-Celeb-1M to pollute them with Gaussian blur noise.
Table~\ref{tab:noise-trainset} demonstrates that our proposed DUL\textsubscript{cls/rgs} perform more robustly on noisy training data.
%------------------------------------------------------------------------

\section{Conclusion}
In this work, we propose two general learning methods to further develop and perfect the \textit{data uncertainty learning} (DUL) for face recognition: DUL\textsubscript{cls} and DUL\textsubscript{rgs}. Both methods give a Gaussian distributional estimation for each face image in the latent space and simultaneously learn identity feature (mean) and uncertainty (variance) of the estimated mean.
Comprehensive experiments demonstrate that our proposed methods perform better than deterministic models on most benchmarks. Additionally, we discuss how the learned \textit{uncertainty} affects the training of model from the perspective of \textit{image noise} by both qualitative analysis and quantitative results.

\section{Acknowledgement}
This paper is supported by the National key R\&D plan of the Ministry of science and technology (Project Name: ``Grid function expansion technology and equipment for community risk prevention'', Project No.2018YFC0809704)
%------------------------------------------------------------------------

{\small
\bibliographystyle{ieee_fullname}
\bibliography{egbib}

\begin{thebibliography}{10}\itemsep=-1pt

\bibitem{alemi1612deep}
Alexander~A Alemi, Ian Fischer, Joshua~V Dillon, and Kevin Murphy.
\newblock Deep variational information bottleneck.
\newblock In {\em Proceedings of the International Conference on Learning
  Representations}, 2017.

\bibitem{bishop1997regression}
Christopher~M Bishop and Cazhaow~S Quazaz.
\newblock Regression with input-dependent noise: A bayesian treatment.
\newblock In {\em Advances in Neural Information Processing Systems}, pages
  347--353, 1997.

\bibitem{blundell2015weight}
Charles Blundell, Julien Cornebise, Koray Kavukcuoglu, and Daan Wierstra.
\newblock Weight uncertainty in neural networks.
\newblock {\em arXiv preprint arXiv:1505.05424}, 2015.

\bibitem{brando2018uncertainty}
Axel Brando, Jose~A Rodr{\'\i}guez-Serrano, Mauricio Ciprian, Roberto Maestre,
  and Jordi Vitri{\`a}.
\newblock Uncertainty modelling in deep networks: Forecasting short and noisy
  series.
\newblock In {\em Joint European Conference on Machine Learning and Knowledge
  Discovery in Databases}, pages 325--340. Springer, 2018.

\bibitem{cao2018vggface2}
Qiong Cao, Li Shen, Weidi Xie, Omkar~M Parkhi, and Andrew Zisserman.
\newblock Vggface2: A dataset for recognising faces across pose and age.
\newblock In {\em 2018 13th IEEE International Conference on Automatic Face \&
  Gesture Recognition (FG 2018)}, pages 67--74. IEEE, 2018.

\bibitem{Choi_2019_ICCV}
Jiwoong Choi, Dayoung Chun, Hyun Kim, and Hyuk-Jae Lee.
\newblock Gaussian yolov3: An accurate and fast object detector using
  localization uncertainty for autonomous driving.
\newblock In {\em The IEEE International Conference on Computer Vision (ICCV)},
  October 2019.

\bibitem{deng2019arcface}
Jiankang Deng, Jia Guo, Niannan Xue, and Stefanos Zafeiriou.
\newblock Arcface: Additive angular margin loss for deep face recognition.
\newblock In {\em Proceedings of the IEEE Conference on Computer Vision and
  Pattern Recognition}, pages 4690--4699, 2019.

\bibitem{der2009aleatory}
Armen Der~Kiureghian and Ove Ditlevsen.
\newblock Aleatory or epistemic? does it matter?
\newblock {\em Structural Safety}, 31(2):105--112, 2009.

\bibitem{faber2005treatment}
Michael~Havbro Faber.
\newblock On the treatment of uncertainties and probabilities in engineering
  decision analysis.
\newblock {\em Journal of Offshore Mechanics and Arctic Engineering},
  127(3):243--248, 2005.

\bibitem{gal2016uncertainty}
Yarin Gal.
\newblock {\em Uncertainty in deep learning}.
\newblock PhD thesis, PhD thesis, University of Cambridge, 2016.

\bibitem{gal2016dropout}
Yarin Gal and Zoubin Ghahramani.
\newblock Dropout as a bayesian approximation: Representing model uncertainty
  in deep learning.
\newblock In {\em international conference on machine learning}, pages
  1050--1059, 2016.

\bibitem{goldberg1998regression}
Paul~W Goldberg, Christopher~KI Williams, and Christopher~M Bishop.
\newblock Regression with input-dependent noise: A gaussian process treatment.
\newblock In {\em Advances in neural information processing systems}, pages
  493--499, 1998.

\bibitem{gong2017capacity}
Sixue Gong, Vishnu~Naresh Boddeti, and Anil~K Jain.
\newblock On the capacity of face representation.
\newblock {\em arXiv preprint arXiv:1709.10433}, 2017.

\bibitem{guo2016ms}
Yandong Guo, Lei Zhang, Yuxiao Hu, Xiaodong He, and Jianfeng Gao.
\newblock Ms-celeb-1m: A dataset and benchmark for large-scale face
  recognition.
\newblock In {\em European Conference on Computer Vision}, pages 87--102.
  Springer, 2016.

\bibitem{he2016deep}
Kaiming He, Xiangyu Zhang, Shaoqing Ren, and Jian Sun.
\newblock Deep residual learning for image recognition.
\newblock In {\em Proceedings of the IEEE conference on computer vision and
  pattern recognition}, pages 770--778, 2016.

\bibitem{hu2018squeeze}
Jie Hu, Li Shen, and Gang Sun.
\newblock Squeeze-and-excitation networks.
\newblock In {\em Proceedings of the IEEE conference on computer vision and
  pattern recognition}, pages 7132--7141, 2018.

\bibitem{hu2019noise}
Wei Hu, Yangyu Huang, Fan Zhang, and Ruirui Li.
\newblock Noise-tolerant paradigm for training face recognition cnns.
\newblock In {\em Proceedings of the IEEE Conference on Computer Vision and
  Pattern Recognition}, pages 11887--11896, 2019.

\bibitem{huang2008labeled}
Gary~B Huang, Marwan Mattar, Tamara Berg, and Eric Learned-Miller.
\newblock Labeled faces in the wild: A database forstudying face recognition in
  unconstrained environments.
\newblock 2008.

\bibitem{isobe2017deep}
Shuya Isobe and Shuichi Arai.
\newblock Deep convolutional encoder-decoder network with model uncertainty for
  semantic segmentation.
\newblock In {\em 2017 IEEE International Conference on INnovations in
  Intelligent SysTems and Applications (INISTA)}, pages 365--370. IEEE, 2017.

\bibitem{kemelmacher2016megaface}
Ira Kemelmacher-Shlizerman, Steven~M Seitz, Daniel Miller, and Evan Brossard.
\newblock The megaface benchmark: 1 million faces for recognition at scale.
\newblock In {\em Proceedings of the IEEE Conference on Computer Vision and
  Pattern Recognition}, pages 4873--4882, 2016.

\bibitem{kendall2015bayesian}
Alex Kendall, Vijay Badrinarayanan, and Roberto Cipolla.
\newblock Bayesian segnet: Model uncertainty in deep convolutional
  encoder-decoder architectures for scene understanding.
\newblock {\em BMVC}, 2015.

\bibitem{kendall2017uncertainties}
Alex Kendall and Yarin Gal.
\newblock What uncertainties do we need in bayesian deep learning for computer
  vision?
\newblock In {\em Advances in neural information processing systems}, pages
  5574--5584, 2017.

\bibitem{khan2019striking}
Salman Khan, Munawar Hayat, Syed~Waqas Zamir, Jianbing Shen, and Ling Shao.
\newblock Striking the right balance with uncertainty.
\newblock In {\em Proceedings of the IEEE Conference on Computer Vision and
  Pattern Recognition}, pages 103--112, 2019.

\bibitem{kingma2013auto}
Diederik~P Kingma and Max Welling.
\newblock Auto-encoding variational bayes.
\newblock {\em ICLR}, 2014.

\bibitem{kraus2019uncertainty}
Florian Kraus and Klaus Dietmayer.
\newblock Uncertainty estimation in one-stage object detection.
\newblock {\em arXiv preprint arXiv:1905.10296}, 2019.

\bibitem{le2005heteroscedastic}
Quoc~V Le, Alex~J Smola, and St{\'e}phane Canu.
\newblock Heteroscedastic gaussian process regression.
\newblock In {\em Proceedings of the 22nd international conference on Machine
  learning}, pages 489--496. ACM, 2005.

\bibitem{liu2017sphereface}
Weiyang Liu, Yandong Wen, Zhiding Yu, Ming Li, Bhiksha Raj, and Le Song.
\newblock Sphereface: Deep hypersphere embedding for face recognition.
\newblock In {\em Proceedings of the IEEE conference on computer vision and
  pattern recognition}, pages 212--220, 2017.

\bibitem{maze2018iarpa}
Brianna Maze, Jocelyn Adams, James~A Duncan, Nathan Kalka, Tim Miller, Charles
  Otto, Anil~K Jain, W~Tyler Niggel, Janet Anderson, Jordan Cheney, et~al.
\newblock Iarpa janus benchmark-c: Face dataset and protocol.
\newblock In {\em 2018 International Conference on Biometrics (ICB)}, pages
  158--165. IEEE, 2018.

\bibitem{ng2014data}
Hong-Wei Ng and Stefan Winkler.
\newblock A data-driven approach to cleaning large face datasets.
\newblock In {\em 2014 IEEE International Conference on Image Processing
  (ICIP)}, pages 343--347. IEEE, 2014.

\bibitem{nix1994estimating}
David~A Nix and Andreas~S Weigend.
\newblock Estimating the mean and variance of the target probability
  distribution.
\newblock In {\em Proceedings of 1994 IEEE International Conference on Neural
  Networks (ICNN'94)}, volume~1, pages 55--60. IEEE, 1994.

\bibitem{pate1996uncertainties}
M~Elisabeth Pat{\'e}-Cornell.
\newblock Uncertainties in risk analysis: Six levels of treatment.
\newblock {\em Reliability Engineering \& System Safety}, 54(2-3):95--111,
  1996.

\bibitem{ranjan2017l2}
Rajeev Ranjan, Carlos~D Castillo, and Rama Chellappa.
\newblock L2-constrained softmax loss for discriminative face verification.
\newblock {\em arXiv preprint arXiv:1703.09507}, 2017.

\bibitem{schroff2015facenet}
Florian Schroff, Dmitry Kalenichenko, and James Philbin.
\newblock Facenet: A unified embedding for face recognition and clustering.
\newblock In {\em Proceedings of the IEEE conference on computer vision and
  pattern recognition}, pages 815--823, 2015.

\bibitem{sengupta2016frontal}
Soumyadip Sengupta, Jun-Cheng Chen, Carlos Castillo, Vishal~M Patel, Rama
  Chellappa, and David~W Jacobs.
\newblock Frontal to profile face verification in the wild.
\newblock In {\em 2016 IEEE Winter Conference on Applications of Computer
  Vision (WACV)}, pages 1--9. IEEE, 2016.

\bibitem{shi2019probabilistic}
Yichun Shi, Anil~K Jain, and Nathan~D Kalka.
\newblock Probabilistic face embeddings.
\newblock In {\em Proceedings of the IEEE International Conference on Computer
  Vision}, 2019.

\bibitem{smith2017cyclical}
Leslie~N Smith.
\newblock Cyclical learning rates for training neural networks.
\newblock In {\em 2017 IEEE Winter Conference on Applications of Computer
  Vision (WACV)}, pages 464--472. IEEE, 2017.

\bibitem{sun2015deeply}
Yi Sun, Xiaogang Wang, and Xiaoou Tang.
\newblock Deeply learned face representations are sparse, selective, and
  robust.
\newblock In {\em Proceedings of the IEEE conference on computer vision and
  pattern recognition}, pages 2892--2900, 2015.

\bibitem{tishby2015deep}
Naftali Tishby and Noga Zaslavsky.
\newblock Deep learning and the information bottleneck principle.
\newblock In {\em 2015 IEEE Information Theory Workshop (ITW)}, pages 1--5.
  IEEE, 2015.

\bibitem{wang2018devil}
Fei Wang, Liren Chen, Cheng Li, Shiyao Huang, Yanjie Chen, Chen Qian, and Chen
  Change~Loy.
\newblock The devil of face recognition is in the noise.
\newblock In {\em Proceedings of the European Conference on Computer Vision
  (ECCV)}, pages 765--780, 2018.

\bibitem{wang2018additive}
Feng Wang, Jian Cheng, Weiyang Liu, and Haijun Liu.
\newblock Additive margin softmax for face verification.
\newblock {\em IEEE Signal Processing Letters}, 25(7):926--930, 2018.

\bibitem{wang2018cosface}
Hao Wang, Yitong Wang, Zheng Zhou, Xing Ji, Dihong Gong, Jingchao Zhou, Zhifeng
  Li, and Wei Liu.
\newblock Cosface: Large margin cosine loss for deep face recognition.
\newblock In {\em Proceedings of the IEEE Conference on Computer Vision and
  Pattern Recognition}, pages 5265--5274, 2018.

\bibitem{wen2016discriminative}
Yandong Wen, Kaipeng Zhang, Zhifeng Li, and Yu Qiao.
\newblock A discriminative feature learning approach for deep face recognition.
\newblock In {\em European conference on computer vision}, pages 499--515.
  Springer, 2016.

\bibitem{wolf2011face}
Lior Wolf, Tal Hassner, and Itay Maoz.
\newblock Face recognition in unconstrained videos with matched background
  similarity.
\newblock In {\em CVPR 2011}.

\bibitem{wu2018light}
Xiang Wu, Ran He, Zhenan Sun, and Tieniu Tan.
\newblock A light cnn for deep face representation with noisy labels.
\newblock {\em IEEE Transactions on Information Forensics and Security},
  13(11):2884--2896, 2018.

\bibitem{xie2018comparator}
Weidi Xie, Li Shen, and Andrew Zisserman.
\newblock Comparator networks.
\newblock In {\em Proceedings of the European Conference on Computer Vision
  (ECCV)}, pages 782--797, 2018.

\bibitem{xie2018multicolumn}
Weidi Xie and Andrew Zisserman.
\newblock Multicolumn networks for face recognition.
\newblock In {\em BMVC}, 2018.

\bibitem{yi2014learning}
Dong Yi, Zhen Lei, Shengcai Liao, and Stan~Z Li.
\newblock Learning face representation from scratch.
\newblock {\em arXiv preprint arXiv:1411.7923}, 2014.

\bibitem{yin2018towards}
Bangjie Yin, Luan Tran, Haoxiang Li, Xiaohui Shen, and Xiaoming Liu.
\newblock Towards interpretable face recognition.
\newblock {\em arXiv preprint arXiv:1805.00611}, 2018.

\bibitem{yin2017multi}
Xi Yin and Xiaoming Liu.
\newblock Multi-task convolutional neural network for pose-invariant face
  recognition.
\newblock {\em IEEE Transactions on Image Processing}, 27(2):964--975, 2017.

\bibitem{yu2019robust}
Tianyuan Yu, Da Li, Yongxin Yang, Timothy~M Hospedales, and Tao Xiang.
\newblock Robust person re-identification by modelling feature uncertainty.
\newblock In {\em Proceedings of the IEEE International Conference on Computer
  Vision}, pages 552--561, 2019.

\bibitem{zafar2019face}
Umara Zafar, Mubeen Ghafoor, Tehseen Zia, Ghufran Ahmed, Ahsan Latif,
  Kaleem~Razzaq Malik, and Abdullahi~Mohamud Sharif.
\newblock Face recognition with bayesian convolutional networks for robust
  surveillance systems.
\newblock {\em EURASIP Journal on Image and Video Processing}, 2019(1):10,
  2019.

\end{thebibliography}
}
\end{document}